# Balancing of Humanoid with Object Mass: Trade-off Analyses and Lifting Control

**Hyunjong Song[1], William Z. Peng[1], and Joo H. Kim[1,2,3]**

**Abstract**

*The demand for humanoid loco-manipulation tasks with an object has recently increased, and most existing control approaches for stability in such tasks rely on heuristics or machine-learning techniques. This study rigorously analyzes and exploits the dynamic effects of the object mass on balance stability. By formulating the object mass parameters in the whole-body dynamics with distributed contact wrenches and centers of pressure at the stance contacts, their nonlinear effects on the system momenta and constraints are quantified. The dynamic models and constraints are incorporated into the construction of the balanced state basin/boundary (BSB), a partition of the center-of-mass state space for a biped system to maintain balance in its desired contacts. The implications of the BSB for prediction and control are highlighted using a humanoid robot and an analytically tractable reduced-order mechanism. The BSBs under different conditions of base of support, actuation capacity, and pose provide systematic analyses of the effects of object mass on the balancing capability of a system. In particular, the trade-off relationships between momentum regulation and limiting factors in balancing are characterized, introducing two key quantities of the object: the critical mass, at which the system's balancing capability is maximum, and the transition mass, which activates different limiting factors. In addition, sufficient conditions for imposing balanced states on a trajectory are established and implemented with BSBs as explicit threshold constraints in the whole-body trajectory optimization for stable object-lifting control of the humanoid, demonstrating the lift-and-hold and lift-and-release tasks with distinct mass properties in simulations and experiments.*

**Keywords**

Balanced state basin; balance stability; critical mass; humanoid robot; lifting control; loco-manipulation; object mass; optimization; transition mass; whole-body dynamics

## 1. Introduction

The increasing demand for humanoid robots in automation has resulted in the recent development of diverse platforms (Ackerman, 2024). Their structural similarities to humans allow for simple integration within human workplaces without modifying existing tools or surroundings to accommodate loco-manipulation tasks. While existing works on the loco-manipulation control of humanoid robots have enabled successful interactions with objects in a wide variety of contexts (Arisumi et al., 2008; Dao et al., 2024; Grey et al., 2014; Han et al., 2020; Li et al., 2023; Murooka, Nozawa, et al., 2017; Murooka, Ueda, et al., 2017; Shigematsu et al., 2018, 2019; Tsuichihara et al., 2024), systematic evaluation of the effects of the object on balance stability and its coupling with task performance have been far less explored. The synthesis of balancing and object manipulation in control remains a significant challenge, especially when considering substantial payloads.

Changes in the mass properties of the total system from manipulated objects or payloads introduce complex interrelations among the available momenta, required joint actuations, and restricted contacts for balance stability, as observed in tightrope walking with a balancing pole (Morasso, 2020; Paoletti and Mahadevan, 2012), load carriage (Mummolo et al., 2016), and lifting tasks (Song et al., 2023a, 2023b). The benefits of additional mass for balancing control or stability have been demonstrated through the use of appendages/tails (Patel and Boje, 2015; Libby et al., 2016; Saab et al., 2018; Shield et al., 2021; Yang et al., 2023), arms (Shield et al., 2021; Khazoom and

[1]Department of Mechanical and Aerospace Engineering, New York University (NYU), New York, USA
[2]NYU Center for Robotics and Embodied Intelligence (CREO), New York, USA
[3]NYU Center for Urban Science and Progress (CUSP), New York, USA

**Corresponding author:**
Joo H. Kim, Department of Mechanical and Aerospace Engineering, New York University, 5 MetroTech Center, #116, Brooklyn, NY 11201, USA
Email: joo.h.kim@nyu.edu

Kim, 2022), balancing poles (Guo et al., 2014), and flywheels (Hofer et al., 2023; Koolen et al., 2012; Lee et al., 2023). However, these investigations are often limited to heuristic findings on the relationship between the additional mass and its benefits. Without a rigorous, quantitative understanding of these effects, no systematic guidelines can be established for the additional mass except that it should not be excessively heavy. Even when control strategies incorporate the mass and other physical properties (dimension, friction coefficient, etc.) of the object, such as in trajectory optimization (Arisumi et al., 2008; Shigematsu et al., 2019), model predictive control (Li et al., 2023), motion planning with graph search (Murooka, Ueda, et al., 2017) and random trees (Grey et al., 2014), and reinforcement learning (Dao et al., 2024), their effects on balance stability are not explicitly quantified. Instead, existing approaches to stable loco-manipulation primarily focus on enhancing task performance or identifying unknown physical properties of the objects (Han et al., 2020; Murooka, Nozawa, et al., 2017; Shigematsu et al., 2018). This knowledge gap motivates the critical contributions of this study—understanding how the mass properties of a manipulated object affect balance stability and performance of a robot during a given whole-body task.

# 2. Background and Contributions

## *2.1 Related Work: Balance Stability in Object Lifting and Load Carriage*

Current stability criteria and control methods were primarily developed for legged gait without a carried object and are unsuitable for general non-locomotion tasks with various object mass properties. Most existing control approaches for load carriage or object manipulation tasks use common stability criteria such as the center-of-mass (CoM) projection (Arisumi et al., 2008; Grey et al., 2014 Dai et al., 2014; Featherstone, 2017), center of pressure (CoP) (Han et al., 2020; Dao et al., 2024), and zero-moment point (ZMP) (Murooka, Ueda, et al., 2017; Shigematsu et al., 2019). These approaches are categorized as reference points (Caron et al., 2017; Goswami, 1999; Kajita et al., 2001; Popovic et al., 2005; Vukobratović and Borovac, 2004). Although they are well established and frequently used for robot gait stability control due to their suitability for direct real-time computation, they are neither necessary nor sufficient to accurately inform balance or fall in general tasks (Peng et al., 2022). In addition, they cannot systematically quantify the comprehensive effects of whole-body dynamics including objects on the balance stability under given system-specific properties and capabilities. These limitations emerge because the conditions of the lifting/load-carriage tasks deviate from the assumptions of reduced-order models of legged gait, such as periodicity and minimal involvement of the upper-body joints.

In principle, other categories of criteria for legged systems, such as limit cycle, state-space partition (Peng et al., 2022), and those implicitly based on dynamic feasibility (Dai et al., 2014; Featherstone, 2017), which have not been actively used for load-carriage tasks, can potentially incorporate the object mass into the total system mass for stability control. Limit cycle approaches (Bhounsule and Zamani, 2017; Hobbelen and Wisse, 2007) evaluate the local stability of periodic motions, employing reduced-order models like the inverted pendulum or compass gait. However, they are unsuitable for inherently aperiodic tasks, such as lifting, as failure to meet limit cycle criteria only indicates the inability to maintain a periodic task, such as steady gait, while balance may be recovered through transition to a non-periodic motion or coming to a stop. Partition-based approaches (Caron et al., 2020; Del Prete et al., 2018; Koolen et al., 2012; Lanari et al., 2014; Pai and Patton, 1997; Posa et al., 2017; Stephens and Atkeson, 2010; Yang et al., 2007; Zaytsev et al., 2018), grounded in the viability kernel concept, define boundaries encompassing stable states or regions to prevent falls. Capture regions and capturability (Koolen et al., 2012) represent computationally tractable partitions that can be used for balance recovery by stepping. This approach aims to predict and control a possible motion for coming to a stop within a certain number of steps without exiting some state- or Euclidean-space partition, as compared with similarly derived reference point criteria, such as capture point (Pratt et al., 2006), divergent component of motion (Englsberger et al., 2015), and their extensions (Griffin et al., 2017; Stephens and Atkeson, 2010; Xinjilefu et al., 2015).

Overall, most of these existing approaches commonly rely on reduced-order models such as the inverted pendulum or their variations induced by simplification and linearized dynamics, which is usually sufficient for online feedback control to prevent falls in certain tasks. On the other hand, due to their lack of whole-body attributes, the corresponding criteria may be insufficient or overly restrictive and are not directly applicable for the whole-body coordination involved in object manipulation or lifting tasks.

### 2.2 Preliminaries: Balanced State Basins with Whole-Body Dynamics

The stabilizing capability of a biped system depends on the system's state and its current/desired contact configurations as well as the system's design and control parameters. The stability is determined with respect to the desired contacts based on the definitions established in the prior works (Peng et al., 2021b, 2022): (a) balanced, if there exists a controlled trajectory starting from the state such that the system does not ever need to alter its contacts, either in single support (SS) or double support (DS); and (b) unbalanced, if all controlled trajectories starting from the state will lead to an inevitable change in the system's contacts. All unbalanced states that are not transitional between SS and DS are falling, which will lead to an undesired contact either before the non-stance foot can reach the desired step (if in SS) or while the feet remain in the existing contacts (if in DS). A stance foot contact is considered maintained with some or no allowance for tipping.

Due to its computationally and analytically tractable dimensions (Mummolo et al., 2015, 2017; Mummolo, Peng, Agarwal, et al., 2018; Mummolo, Peng, Gonzalez, et al., 2018), the CoM-state (position and velocity) space is suitable for evaluation and high-level control and planning. Within the CoM-state space, the balanced state boundary (BSB) is the set of all states with maximum allowable CoM velocity perturbations in a given direction, under which a biped system can maintain its desired contact. Accordingly, the aforementioned definitions are incorporated into the mathematical construction of BSB formed by solving a series of optimization problems. The balanced state basin (also abbreviated as BSB as will be clear within the context), a partition of the system's state space defined by the boundaries, represents the set of all reachable balanced states for the desired contacts without *a priori* assumption of a specific control law or controller properties. A state outside of BSB represents the sufficient condition for unbalanced, leading to an unavoidable change in contact.

While the BSBs are mapped into the system's CoM-state space for observability with a reduced dimension, the construction of BSB is based on the whole-body dynamics. Unlike most existing state-space partition approaches relying on reduced-order models (Englsberger et al., 2015; Hopkins et al., 2015; Koolen et al., 2012; Stephens and Atkeson, 2010), the models of whole-body dynamics are critical in several aspects of this work. In previous studies (Peng et al., 2021b, 2022; Song et al., 2024), the significance of a robot's whole-body model when evaluating stability was proven by its notable differences not only against those derived from typical reduced-order models but also across those for different balancing strategies and contacts within the whole-body scheme. The feasibility of BSBs was validated in evaluations with existing gait controls (Mummolo, Peng, Agarwal, et al., 2018; Peng et al., 2021b, 2022) and as a threshold trigger for discrete strategies in a push-recovery control (Song et al., 2024; Bodmer et al., 2026).

### 2.3 Contributions and Organization

Whole-body properties and dynamics are critical when the system and task do not conform to common assumptions of periodic gait or reduced order. The stability of a humanoid robot during a loco-manipulation task depends on its properties (e.g., actuation capacities) and contact conditions (e.g., step lengths), which cannot be represented by reduced-order models, and its upper extremities, whose dynamics are often neglected in controls due to their low mass relative to that of the total system. Coordination between the upper extremities, which handle the object, and the rest of the body is therefore essential for concurrent balancing and task performance. Furthermore, existing works on loco-manipulation mostly focus on control without comprehensive understanding or prediction of the quantitative effects of the object mass on balance stability. This study tackles these limitations through a systematic approach based on full-order, nonlinear dynamics of whole-body models to rigorously analyze and exploit the effects of the object mass on balance stability.

Specific contributions of this study are as follows:

(a) Formulation. While the mass of an object a robot can handle without falling is limited by the constraints inherent in the system and contacts, it can also promote balancing through the increased available momenta. The theoretical foundation of this nonlinear trade-off relationship, which was not rigorously explored in the literature, is established by formulating the object mass parameters in the whole-body dynamics with distributed contact wrenches and CoPs at the legged stance contacts. This allows the systematic quantification of the dynamic effects of the object mass on balancing and the associated active constraints that determine the limiting factors.

(b) Prediction. By incorporating the dynamic models and constraints into the extended construction of BSBs, the nonlinear effects of the object mass on balancing are systematically analyzed. The BSBs with various object masses under different conditions of the robot's base of support (BoS), actuation capacity, and pose provide quantitative prediction and evaluation of its balancing capability. In particular, the trade-off relationships between momentum regulation and limiting factors in balancing are characterized, introducing two key quantities of the object: the critical mass, at which the system's balancing capability is maximum, and the transition mass, which activates different limiting factors.

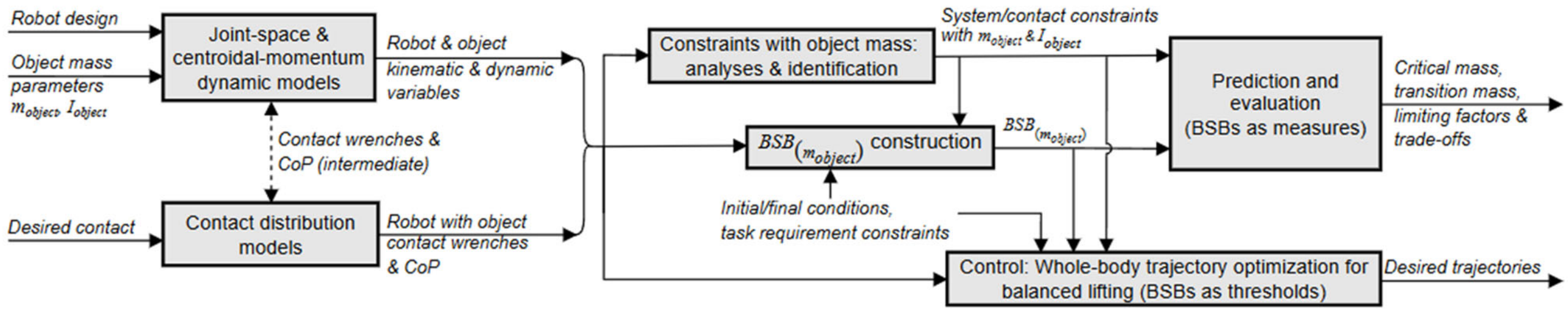

**Figure 1.** Schematic illustration of the overall approach

(c) Control. Sufficient conditions for imposing balanced states on a trajectory for fall prevention are established with respect to collocation points and change in the mass properties. These conditions inform the use of the BSBs as explicit threshold constraints in the whole-body trajectory optimization, and are demonstrated for stable object-lifting control of the humanoid robot tasks without and with change in the mass (lift-and-hold and lift-and-release).

The next section formulates and analyzes the dynamics, constraints, and trade-off relationships. Section 4 describes the modeling of the two systems used in this study—a humanoid robot as the main platform for implementation and a reduced-order mechanism as an analytically tractable complement. Section 5 extends the construction of BSB. Section 6 establishes and implements a trajectory optimization framework for balanced lifting control. Sections 7 and 8 present the analytical, simulation, and experimental results and their validations.

## 3. Object Mass in Balancing: Dynamics and Trade-off Analyses

Consider a robot in a legged stance with an object at its end-effector. The joint-space dynamics are used to model the joint kinematics and actuations, the Cartesian-space centroidal dynamics to model the CoM states and system momenta, and both to formulate the contact wrench and the CoP at each stance contact interaction with the environment (e.g., the ground). The mass parameters of the object are incorporated into these dynamic models and the associated constraints for quantitative evaluations of their nonlinear effects and trade-offs in balancing (Figure 1).

### *3.1 Joint-Space and Centroidal-Momentum Dynamics with Object Mass Parameters*

For an $n$-degree-of-freedom (DoF) robot with an object rigidly attached to its end-effector(s) (e.g., hand), the joint-space dynamics can be described using the general formulation (Mummolo, Peng, Gonzalez, et al., 2018; Peng et al., 2022):

$$\boldsymbol{\tau}_{gen} = M\left(\mathbf{q}_{gen}\right)\ddot{\mathbf{q}}_{gen} + \mathbf{h}\left(\mathbf{q}_{gen}, \dot{\mathbf{q}}_{gen}\right) + \mathbf{c}^{stance}\left(\mathbf{q}_{gen}, \cdot\right) + \mathbf{c}^{other}\left(\mathbf{q}_{gen}, \cdot\right) \tag{1}$$

where a dot indicates time-derivative. The vector of generalized coordinates $\mathbf{q}_{gen} = [\mathbf{q}^T \quad \mathbf{q}_{base}^T]^T \in \mathbb{R}^{n+6}$ includes the joint variables $\mathbf{q} \in \mathbb{R}^n$ and the global position and orientation $\mathbf{q}_{base} \in \mathbb{R}^6$ of the unactuated floating base of the system with respect to a three-dimensional global reference frame. The corresponding generalized torques $\boldsymbol{\tau}_{gen} = [\boldsymbol{\tau}^T \quad \boldsymbol{\tau}_{base}^T]^T \in \mathbb{R}^{n+6}$ include the joint actuation $\boldsymbol{\tau} \in \mathbb{R}^n$ and the zero free-floating-base forces/moments, $\boldsymbol{\tau}_{base} = \mathbf{0} \in \mathbb{R}^6$. The mass-inertia matrix $M(\mathbf{q}_{gen}) \in \mathbb{R}^{(n+6)\times(n+6)}$ and the vector of generalized Coriolis-centrifugal forces and gravitational torques $\mathbf{h}(\mathbf{q}_{gen}, \dot{\mathbf{q}}_{gen}) \in \mathbb{R}^{n+6}$ explicitly include the mass parameters of the object—i.e., mass $m_{object}$ and inertia tensor $I_{object}$ with respect to the nonrotating centroidal local frame attached to the object—as well as those of the robot. The terms $\mathbf{c}^{stance}(\mathbf{q}_{gen}, \cdot) \in \mathbb{R}^{n+6}$ and $\mathbf{c}^{other}(\mathbf{q}_{gen}, \cdot) \in \mathbb{R}^{n+6}$ are the generalized torques of the Jacobian-mapped contact wrenches from the environment through the desired stance contacts and any other undesired body parts, respectively. The resultant contact wrench $\{\mathbf{F}_p, \mathbf{M}_p\}$ is applied on the $p$-th ($p = 1, 2$ for a biped system) stance contact position $\mathbf{r}_p$ with respect to the global frame, where $\mathbf{F}_p$ is the force and $\mathbf{M}_p$ is the moment applied by the environment. Any contact wrench $\{\mathbf{F}_{other}, \mathbf{M}_{other}\}$ applied on the other body parts by the environment would result from a falling state.

In the Cartesian space, the linear momentum of the robot with mass $m_{robot}$ is $\mathbf{L}_{robot} = m_{robot}\dot{\bar{\mathbf{r}}}_{robot}$, and of the object is $\mathbf{L}_{object} = m_{object}\dot{\bar{\mathbf{r}}}_{object}$, where $\bar{\mathbf{r}}$ is the CoM position of a system with respect to the global frame, indicated with the subscripts *robot* and *object* if it is not for the total system. Then the resultant applied force includes the additionally available linear momentum rate due to the object mass as follows:

$$\sum_p \mathbf{F}_p + (m_{robot} + m_{object})\mathbf{g} = \dot{\mathbf{L}}_{robot} + \dot{\mathbf{L}}_{object} \tag{2}$$

Here, $\mathbf{g} = [0\ 0\ -g]^T$ is the gravitational acceleration vector with $g = 9.8062$ m/s$^2$.

The centroidal angular momentum of the robot is $\mathbf{H}_{c,robot} = \sum_k (\boldsymbol{\rho}_k \times m_k \dot{\boldsymbol{\rho}}_k + I_k \boldsymbol{\omega}_k)$, where $k$ is the link index, $\boldsymbol{\rho}_k$ is the link CoM position vector relative to the total (i.e., for the combined system of the robot and the object) CoM, $\boldsymbol{\omega}_k$ is the link absolute angular velocity, and $I_k$ is the time-varying link inertia tensor with respect to the nonrotating link-centroidal local frame. Similarly, $\mathbf{H}_{c,object} = \boldsymbol{\rho}_{object} \times m_{object}\dot{\boldsymbol{\rho}}_{object} + I_{object}\boldsymbol{\omega}_{object}$ for the object. Then the resultant applied moment about the total CoM includes the additionally available angular momentum rate due to the object mass parameters as follows:

$$\sum_p \{\mathbf{M}_p + (\mathbf{r}_p - \bar{\mathbf{r}}) \times \mathbf{F}_p\} = \dot{\mathbf{H}}_{c,robot} + \dot{\mathbf{H}}_{c,object} \tag{3}$$

### *3.2 Transfer and Distribution of Object Dynamics to Contact Wrenches and CoPs*

Under a rigid-body assumption, the distributed reaction forces at stance contact surfaces from the environment can be represented by CoPs and the resultant contact wrenches applied at their respective reference points. For DS, the mechanical indeterminacy in the contact wrench and CoP distributions across both contacts under a no-slip condition is resolved by introducing time-varying parameters $\boldsymbol{\alpha}$ (Peng et al., 2022) that are to be determined for one (either $p$ = 1 or 2) of the stance contacts:

$$\mathbf{F}_p\text{:}\quad \mathbf{F}_p \bullet \hat{\mathbf{n}}_p = \alpha_1\{\dot{\mathbf{L}}_{robot} + \dot{\mathbf{L}}_{object} - (m_{robot} + m_{object})\mathbf{g}\} \bullet \hat{\mathbf{n}}_p\ ;\quad \mathbf{F}_p \bullet \hat{\mathbf{t}}_p(\alpha_3) = (2\alpha_2 - 1)\mu \| \mathbf{F}_p \bullet \hat{\mathbf{n}}_p \| \tag{4}$$

$$\mathbf{M}_p\text{:}\quad \mathbf{M}_p \bullet \hat{\mathbf{n}}_p = (2\alpha_4 - 1)M_f^{\max}(\cdot)\ ;\quad \mathbf{M}_p - (\mathbf{M}_p \bullet \hat{\mathbf{n}}_p)\hat{\mathbf{n}}_p = {}^p\mathbf{r}_{CoP}(\alpha_5, \alpha_6) \times (\mathbf{F}_p \bullet \hat{\mathbf{n}}_p)\hat{\mathbf{n}}_p \tag{5}$$

where $\boldsymbol{\alpha} \in \mathbb{R}^6$ is a non-dimensional vector with each element normalized within [0, 1]; $\hat{\mathbf{n}}_p$ and $\hat{\mathbf{t}}_p$ are the unit vectors that are normal and tangential in the direction of the friction force, respectively, to the $p$-th contact surface; $\mu$ is the coefficient of static friction; $M_f^{\max}(\cdot)$ is the maximum static frictional moment normal to the contact surface, which is a function of normal contact force distribution (including the dynamic effects of the object's mass), geometry, area, and $\mu$ (Beer et al., 2009); and ${}^p\mathbf{r}_{CoP}$ is the CoP position relative to the local frame of the $p$-th contact link projected onto the contact surface.

The normal forces are distributed by the ratio $\alpha_1$. The transformations $2\alpha_2 - 1 \in [-1,1]$ and $2\alpha_4 - 1 \in [-1,1]$ determine the bidirectional friction force and moment, respectively, within their maximum static values. Within the contact surface, $\hat{\mathbf{t}}_p$ is parameterized by an orientation variable $\alpha_3$ (e.g., angle) and ${}^p\mathbf{r}_{CoP}$ by the variables $\alpha_5$ and $\alpha_6$ (e.g., local coordinates) that represent a point on the BoS surface. These parameterizations for one of the stance contacts are coupled with the joint-space and centroidal-momentum dynamics to determine the contact wrenches and CoPs in both stance contacts.

The time-varying contact wrenches are mapped to the joint space as the generalized torque term $\mathbf{c}^{stance}$ for the desired stance contacts. As results, the joint-space dynamics formulation incorporates the effects of the object's mass parameters explicitly through its mass-inertia and Coriolis-centrifugal-gravity terms and implicitly through the generalized contact torque term. Overall, the combined models of the joint-space and centroidal-momentum dynamics and the formulations of the contact wrench and CoP distribution provide the quantification of the dynamic effects of the object's mass, which are transferred to the robot through the joint actuations and linear/angular momenta, and eventually to the stance contact surfaces (Figure 1). Since the contact wrenches and CoPs at the desired contacts are the only external force interactions between the system and the environment other than the gravity, they have the operational-space control authority to regulate the robot's balancing.

### *3.3 Trade-off of Object Mass in Balancing: System/Contact Constraints and Limiting Factors*

According to the dynamic models, for given joint states, the object's mass parameter values are positively correlated with the required joint actuation torques and the available linear and angular momenta of the system, which positively correlate with BSB through the available regulation of the contact wrenches and CoPs. Therefore, a larger mass of the object always increases the balancing capabilities of the robot, absent the constraints imposed by its system parameters and contact conditions with the environment (Mummolo, Peng, Gonzalez, et al., 2018; Peng et al., 2021b). Due to their coupling with the dynamic models, the system and contact constraints, as determined by their lower (*LB*) and upper bounds (*UB*), become the limiting factors for the robot's balancing capabilities with respect to the object's mass.

*System Joint Kinematics and Actuations.* The position and velocity kinematics of each joint of the robot and the associated actuation are bounded by the system design and properties (including any requirements for self-collision avoidance):

$$\begin{bmatrix} \mathbf{q} \\ \dot{\mathbf{q}} \\ \boldsymbol{\tau}(\mathbf{q},\dot{\mathbf{q}}) \end{bmatrix}^{LB} \leq \begin{bmatrix} \mathbf{q} \\ \dot{\mathbf{q}} \\ \boldsymbol{\tau} \end{bmatrix} \leq \begin{bmatrix} \mathbf{q} \\ \dot{\mathbf{q}} \\ \boldsymbol{\tau}(\mathbf{q},\dot{\mathbf{q}}) \end{bmatrix}^{UB} \quad (6)$$

In the joint-space dynamics, the mass-inertia matrix and the Coriolis-centrifugal-gravitational vector are linear with respect to the system's mass parameters (Sciavicco and Siciliano, 2012; Spong et al., 2020), including the additional mass of the object, with a regressor matrix for given joint kinematics. Due to this, along with their positive correlations with the resultant generalized contact wrenches, the object's mass parameter values that can be handled by the robot are limited by the bounded joint actuation torques. The object's mass parameters beyond this limit would result in the reduced balancing capabilities of the robot. In addition, the limited availability in the linear/angular momenta of the robot and the object due to the bounded kinematics and actuations restricts the robot's active regulations of the contact wrenches and CoPs at the stance contact interactions.

*Contact BoS and Unilateral Normal Reaction.* For each $p$-th stance, the contact reaction force is unilateral in the positive direction of the surface normal and the CoP is contained within the BoS under a constant-contact (i.e., no tipping or rolling) condition:

$$\mathbf{F}_p \bullet \hat{\mathbf{n}}_p \geq 0\,; \quad {}^{p}\mathbf{r}_{CoP} \in BoS_p \quad (7)$$

with $BoS_p$ the set of position vectors of the points forming the $p$-th stance contact BoS. The CoP constraint is implicitly imposed for one of the stance contacts through the contact wrench distribution formulations by the bounds on $\alpha_5$ and $\alpha_6$. As a result, the additionally available momenta of the system due to the object's mass are limited. Beyond these limits, the resulting foot tipping may lead the system to either stay balanced or fall, depending on the given state of the system (Peng et al., 2021a).

*Contact Friction Cone.* Under a no-slip condition, each $p$-th stance contact is subject to the static friction cone:

$$|\,\mathbf{F}_p \bullet \hat{\mathbf{t}}_p\,| \leq \mu(\mathbf{F}_p \bullet \hat{\mathbf{n}}_p) \quad (8)$$

According to the formulations for the contact wrench and CoP distribution, which implicitly impose this constraint for one of the stance contacts by the bounds on $\alpha_2$, the positive correlations between the object's mass parameters and $\mathbf{F}_p$ affect both sides of the inequality. As the object's mass parameter values increase, the maximum static friction force will generally increase, but the friction force may also increase due to the increased system momenta.

The constraint(s) that act as the limiting factors for the BSBs depends on the relative values of the constraint function and bounds. Overall, the coupled system of the dynamic models and the system/contact constraints formalizes the trade-offs between the available momenta and the limiting factors with respect to the object mass in balancing (Figure 1).

## 4. Systems and Models

Two systems are used for validations and demonstrations: a whole-body humanoid robot and a 2-DoF reduced-order mechanism on a support link (Figure 2). The reduced-order mechanism is an analytically tractable complement to the humanoid robot, with similar system properties, allowing for independent evaluations of various conditions without the complexity from large DoFs and multiple contacts. Planar models of these systems are implemented in the global coordinate frame $\{X, Y, Z\}$, in which the sagittal $XY$-plane is the plane of interest with the level ground surface at $Y = 0$, and the rotations are in the out-of-plane $Z$-component. While these conditions provide analytical simplicity and computational tractability with visualizable results in this study, the proposed dynamics formulations and constraint analyses can be implemented for general 3D balancing in any directions.

### *4.1 Humanoid Robot*

A ROBOTIS-OP3 humanoid robot (ROBOTIS Co., Ltd.; Gao et al., 2020; Haarnoja et al., 2024) serves as the whole-body platform, which has 20 DoFs: two for the neck, three for each arm, and six for each leg (Figure 2). The robot mass $m_{robot} = 3.5$ kg includes 1.74 kg for torso, 0.23 kg for each arm, and 0.59 kg for each leg. The upright standing total height is 0.51 m and $fl = 0.127$ m.

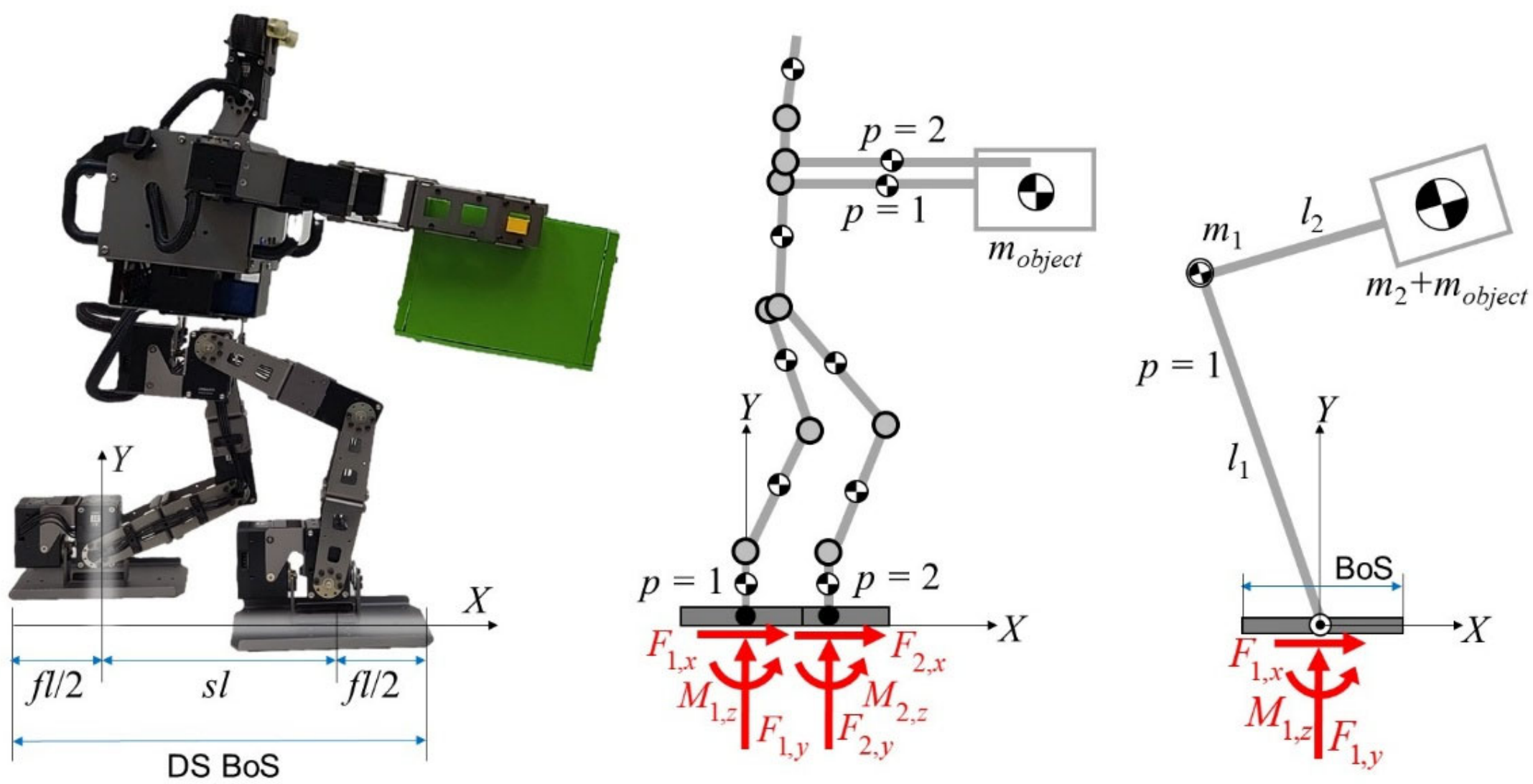

**Figure 2.** The ROBOTIS-OP3 humanoid robot holding an object in DS (left), its sagittal-plane model with ground reaction forces and moments (center), and the reduced-order mechanism on a support link (right). For the humanoid robot, each side of the limbs is indexed with $p = 1, 2$.

The total DoFs of the whole-body model in the sagittal plane consist of 12 joint variables including the nine in-plane rotational joints of the robot (Figure 2) and the three DoFs (two translational and one rotational) for the floating base at the torso. The mass, CoM, and inertia tensor of each link are based on the manufacturer data (ROBOTIS-GIT/ROBOTIS-OP3-Common, n.d.). For simplicity, the object is included in the model as a point mass $m_{object}$ that is equally distributed and added to the distal end of each arm. The joint variable limits were obtained from the in-plane ranges of motion measured by the servomotor encoders. A hexagonal domain, similar to those from prior works (Peng et al., 2021b, 2022), is used for the joint torque-velocity limits bounded by the servomotor's no-load speed of 4.82 rad/s and stall torque limit $\tau_{ref}^{UB} = 3.56$ Nm (ROBOTIS e-Manual). The dimension of its BoS during DS is the sum of $fl$ and the step length $sl$. The position $\mathbf{r}_p$, on which the contact wrench is applied, is at the center of each $p$-th stance foot. Without loss of generality, the origin of the global frame is at the center of the $p = 1$ stance foot such that $(r_{1,x}, r_{1,y}) = (0, 0)$ and $(r_{2,x}, r_{2,y}) = (sl, 0)$ for DS on level ground.

## *4.2 Reduced-Order Mechanism and Limiting Factor Analyses*

The 2-DoF system on a massless support link (with only $p = 1$) serves as a reduced-order representation of the humanoid robot. The concentrated masses $m_1$ and $m_2$ and the lengths $l_1$ and $l_2$ of the links are assigned to represent a lifting task similar to that of the humanoid robot (Figure 2). The mechanism has no joint limits and shares the symmetric constant velocity and torque limits $\tau_{ref}^{UB}$ of the robot as references. The support link length is the BoS dimension and the humanoid robot's foot length $fl$ serves as its reference value $BoS_{ref}$. The origin of the global frame is at the center of the support link such that $(r_{1,x}, r_{1,y}) = (0, 0)$ is where the contact wrench is applied. Among the system/contact constraints as the potential limiting factors for balancing, the torque limit for the actuation capacity and the BoS dimension for the CoP bounds are evaluated with their variations due to the availability of their direct cause-and-effect relationship analyses.

In this study, at the time instant when a perturbation is applied in the $X$-direction, the initial vertical CoM velocity and acceleration are constrained to be zero. While the approach is applicable for any initial pose for a given $m_{object}$, for consistency in the analyses, it is set for the CoM $X$-position at the center of the support link ($\bar{r}_x(0) = 0$) and the horizontal second link. Then, the CoP $X$-position ${}^{1}r_{CoP,x} = \tau_1 / (m_1 + m_2 + m_{object})g$ with respect to the local frame at the center of the support link is bounded by its BoS $X$-dimension as $-0.5BoS_x \le {}^1r_{CoP,x} \le 0.5BoS_x$ for maintaining a full contact, which can be written in terms of the joint torque as follows:

$$|\tau_1| \le 0.5BoS_x(m_1 + m_2 + m_{object})g \tag{9}$$

Since the required joint torque has a positive correlation with $m_{object}$ for a given pose and is also bounded by $|\tau_1| \le \tau^{UB}$, the active/inactive statuses of the CoP and torque constraints depend on the relative magnitudes of their bounds. For given bounds $\tau^{UB}$ and $BoS_x$, the available joint actuation to regulate the system momenta for balancing is lower-bounded by the torques required for the initial pose and upper-bounded by either the torque limit or the torques required to satisfy the CoP constraint, depending on $m_{object}$. Therefore, the object mass corresponding to the identical bounds for both constraints (i.e., $0.5BoS_x(m_1 + m_2 + m_{object})g = \tau^{UB}$ ) while satisfying the constraint on the torque required for the initial pose (i.e., $\tau^{UB} \ge (m_2 + m_{object})gl_2$ for the second joint) serves as the transition mass that switches active and inactive statuses of the constraints:

$$m_{obj,trans} = \frac{2\tau^{UB}}{BoS_x \cdot g} - m_1 - m_2 \le \frac{\tau^{UB}}{l_2 \cdot g} - m_2 \tag{10}$$

The limiting factors for balancing according to the constraint statuses depend on $m_{object}$ relative to $m_{obj,trans}$:

**1**. $m_{object} < m_{obj,trans}$ (or $0.5BoS_x(m_1 + m_2 + m_{object})g < \tau^{UB}$). If the CoP constraint is active, the minimal joint torque required is $\tau_1|_{m_{object}=0} = 0.5BoS_x(m_1 + m_2)g$ when $m_{object} = 0$. Thus, a torque limit with $\tau^{UB} > 0.5BoS_x(m_1 + m_2)g$ is required for the CoP constraint to be active. This implies that, with these sufficiently large torque limits, its constraint is inactive; in other words, the actuation capacity is not fully exploitable in balancing. Therefore, the BoS dimension is the limiting factor.

**2**. $m_{object} = m_{obj,trans}$ (or $0.5BoS_x(m_1 + m_2 + m_{object})g = \tau^{UB}$). Both constraints have identical active/inactive statuses, which are switched at this object mass value, and simultaneously serve as the limiting factors for balancing. For a given actuation capacity, the transition mass decreases as the BoS dimension increases. For a given BoS dimension, the transition mass increases as the actuation capacity increases.

**3**. $m_{object} > m_{obj,trans}$ (or $0.5BoS_x(m_1 + m_2 + m_{object})g > \tau^{UB}$). An active status of the torque limit constraint indicates an inactive status of the CoP constraint. This implies that the system's CoP cannot utilize the full range of the BoS dimension, leading to a decrease in the balancing capability as $m_{object}$ increases. Therefore, the actuation capacity is the limiting factor.

## 5. Balanced State Boundaries for Humanoid with Object Mass

An optimization-based framework from prior works (Peng et al., 2022; Song et al., 2024) is extended here to construct BSBs in the CoM-state space of the humanoid robot with an object mass. Each BSB is computed as the solutions to a series of constrained nonlinear optimization problems over a discretized domain of whole-body poses. For each pose at the initial time $t = 0$ when a perturbation is applied, the magnitude of its CoM velocity along a specified direction is maximized, which represents the maximum allowable CoM velocity perturbation that a system with a given $m_{object}$ can sustain without changing its desired current contacts. Any perturbation applied at another position in the system can be transported to the CoM as an equivalent screw. With the fore-aft $X$-axis in the sagittal plane as the perturbation direction of interest, the magnitude $\dot{r}_x(0)$ of the CoM $X$-velocity is to be maximized:

$$\max_{\mathbf{v},\boldsymbol{\alpha}} \dot{r}_x(0) \text{ subject to } \mathbf{b}_{BSB}^{LB} \le \mathbf{b}_{BSB}(\mathbf{v},\boldsymbol{\alpha}) \le \mathbf{b}_{BSB}^{UB} \tag{11}$$

where $\mathbf{b}_{BSB}$ is the vector of constraint functions required for the BSB construction. The optimization variables are the third-order B-spline control vertices $\mathbf{v}$ (Kim, 2011), which parameterize the joint kinematics $\mathbf{q}$, $\dot{\mathbf{q}}$, and $\ddot{\mathbf{q}}$, and the non-dimensional variables $\boldsymbol{\alpha}$ for the contact wrench distribution, which are mapped for the 2D sagittal plane and time-discretized. The previously-derived system and contact constraints are imposed along with the joint-space and centroidal-momentum dynamics over the entire solution time interval [0, $T$], where $T$ is a sufficient terminal time. Additional constraints are imposed for the initial/final conditions, maintaining balance, and task requirements (Figure 1), as below.

Each sampled whole-body pose $\mathbf{q}^{sampled}$ is imposed as an initial condition, while the zero initial CoM $Y$-velocity

$\dot{\bar{r}}_y(0)$ and acceleration $\ddot{\bar{r}}_y(0)$ confine the perturbation to the $X$-direction:

$$\mathbf{q}(0) = \mathbf{q}^{sampled}; \;\; \dot{\bar{r}}_y(0) = \ddot{\bar{r}}_y(0) = 0 \tag{12}$$

The conditions of the balanced state require no motion of the current contact points relative to the ground such as slipping or tipping. For DS, both feet ($p$ = 1, 2) of the humanoid robot are fixed to the ground at the *desired* positions at all times ($\forall t \in [0, T]$):

$$\mathbf{r}_p(t) = \mathbf{r}_p^{desired}; \;\; \Phi_{p,foot}(t) = 0 \tag{13}$$

where $\Phi_{p,foot}$ is the angle of the $p$-th foot sole from the horizontal in a counterclockwise direction and is zero for the full foot contact with level ground. In addition, the contact wrenches $\{\mathbf{F}_{other}, \mathbf{M}_{other}\}$ associated with any undesired (other than foot) contacts, which would result from a falling state, are required to be null at all times:

$$\{\mathbf{F}_{other}, \mathbf{M}_{other}\} = \{\mathbf{0}, \mathbf{0}\} \tag{14}$$

To ensure that the system maintains its contacts after the solution time interval, the CoM static equilibrium within a specified CoM $X$-position range that is contained within the BoS is imposed as the final conditions:

$$\bar{r}_x^{LB} \le \bar{r}_x(T) \le \bar{r}_x^{UB}; \;\; \dot{\bar{\mathbf{r}}}(T) = \ddot{\bar{\mathbf{r}}}(T) = \mathbf{0} \tag{15}$$

As task requirements for the humanoid robot to hold the object with sagittal rotation and to avoid self-collision, both arm configurations are identical and their orientations are restricted within a specified range throughout the solution time interval ($\forall t \in [0, T]$):

$$q_{1,arm}(t) = q_{2,arm}(t); \;\; \Phi_{arm}^{LB} \le \Phi_{1,arm}(t) = \Phi_{2,arm}(t) \le \Phi_{arm}^{UB} \tag{16}$$

where $q_{p,arm}$ is the shoulder joint pitch angle and $\Phi_{p,arm}$ is the angle of the arm ($p$ = 1, 2) from the horizontal.

The problem is solved by a sequential quadratic programming algorithm (Gill et al., 2002) with a direct collocation scheme. The algorithm convergence is ensured by incorporating function continuity/boundedness (Arora, 2017), analytical gradients, and feasible initial solutions (Peng et al., 2021b). Numerical experiments inform the selections of: initial solutions for their sensitivity toward optimality; B-spline parameters and time-discretization for accuracy in the collocation scheme; and pose domain discretization for a sufficient sampling. The resulting CoM state points form the boundaries of BSB, equivalent to 0-step capturability (Peng et al., 2022), which represents the balancing capability as a property or the threshold for control of the robot with the object. The framework can be extended to include step-based capturability (Peng et al., 2022; Song et al., 2024; Bodmer et al., 2026). For notational simplicity under the assumption that all other parameters and current pose are fixed for the humanoid robot within a context, the BSB as a basin or a boundary corresponding to a given $m_{object}$ is denoted as $BSB_{(m_{object})}$ when needed.

The object mass parameters incorporated into the dynamic models and the associated constraints induce nonlinear effects and trade-offs in the resulting BSBs. While additional mass can enhance balancing capability by the increased available linear and angular momenta through the contact wrench and CoP regulations, this enhancement is simultaneously limited by the system and contact constraints. This results in the critical mass of the object, at which the system's balancing capability reaches its maximum, as well as the transition mass, which activates different limiting factors. The BSB framework enables quantitative prediction of this trade-off behavior and the associated critical and transition masses (Figure 1).

## 6. Whole-body Trajectory Optimization for Balanced Lifting Control

The use of BSBs as stability thresholds is analyzed and demonstrated by imposing them as explicit constraints in the whole-body trajectory optimization problems for robot control (Figure 1). Two object lifting tasks from an initial squat pose are considered for proofs of concept: lift-and-hold, where the robot holds the object throughout the task; and lift-and-release, where the robot instantaneously releases the object once a standing pose is reached before returning to the initial squat pose, representing a repetitive operation in common practice (Figure 3). These two tasks demonstrate distinct characteristics in balancing with respect to mass properties.

### *6.1 Sufficient Conditions for Imposing Balanced States on a Trajectory*

Sufficient conditions for imposing balanced states on a trajectory for fall prevention are established with respect to collocation points and change in the total system's mass properties. These conditions inform the use of the BSBs as explicit threshold constraints in the trajectory optimization for the stable humanoid robot control.

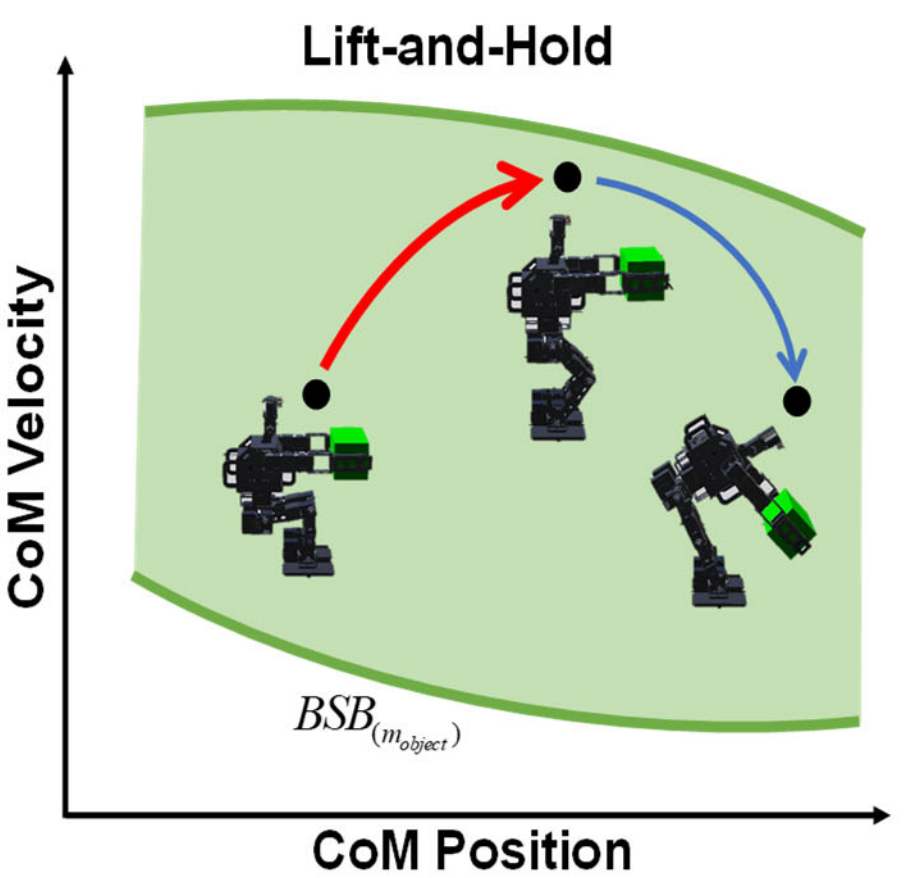

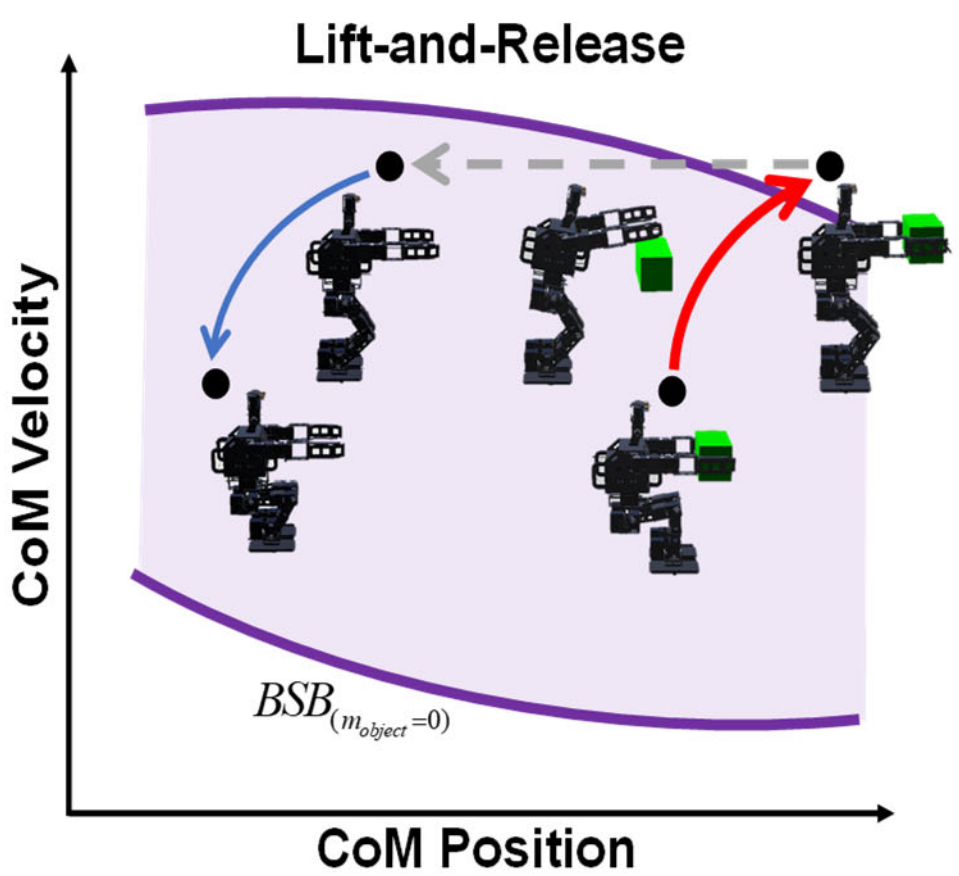

**Figure 3.** Schematic illustration of the proposed trajectory optimization for the lift-and-hold and lift-and-release tasks: squat-to-stand (thick arrow) and stabilization (thin arrow) phases. The release of the object results in an instantaneous shift in the total CoM position (dashed arrow).

In general, if a given system has invariant mass properties and remains in its unchanged desired contact during a time span, a current balanced state indicates that all previous states in that time span are also balanced. In this case, imposing the BSB constraint on the current state is sufficient to ensure that all previous states in the given desired contact are balanced throughout the trajectory. This can be shown by contradiction: If there exists any previous state in the trajectory that exits the corresponding BSB, then, according to the proposed definitions, that state will lead to an unavoidable change in contact, resulting in an eventual fall or stepping. This violates the BSB constraints with respect to the given desired contact for the subsequent and current states. Therefore, in the lift-and-hold task, the total CoM state at the end of lifting should be within the BSB with the object mass (Figure 3, left).

If a system's balancing capability (i.e., BSB) in a desired contact is changed at a current instant due to any change in the system's properties (e.g., mass or actuation capacity), then the above statement can be extended to include any previous states that may be temporarily unbalanced while remaining in the given desired contact. Unless any previous unbalanced state leads to a change of contact (resulting in a fall or stepping) prior to the current time, the balance stability of the current state depends on the changed balancing capability. In this case, imposing the BSB constraint with the changed system properties on the current state will be sufficient to ensure that all previous states remain in the given desired contact throughout the trajectory. Therefore, in the lift-and-release task, the CoM state of the robot without the object should be within the BSB for zero object mass (Figure 3, right). This ensures that the robot can achieve balance after instantaneously releasing the object, while allowing the total CoM state to exit the BSB with the object mass at the time of release.

## *6.2 Optimization Formulations*

All lifting tasks are in DS with a given $sl$ on level ground within the sagittal plane. In each task with a given $m_{object}$, the lifting trajectory is computed in two consecutive phases, transitioning from an initial squat to an intermediate standing pose at $t = T_1$ and the subsequent stabilization to achieve final static balance at a sufficient time $t = T_2$. Each of these phases is formulated in a direct collocation scheme and their solutions are joined to obtain the entire task trajectory. Like in the BSB construction, the problems are solved by a sequential quadratic programming algorithm with the convergence requirements.

For the squat-to-stand phase trajectories during $[0, T_1]$, a measure of control effort (Remy et al., 2012) is minimized, while the stabilization phase trajectories during $[T_1, T_2]$ are obtained as feasible solutions for achieving final static balance:

$$\min_{\mathbf{v},\boldsymbol{\alpha}} \int_0^{T_1} \left\|\boldsymbol{\tau}(t)\right\|^2 dt$$
$$\text{subject to} \quad \mathbf{b}_{lifting}^{LB} \le \mathbf{b}_{lifting}(\mathbf{v},\boldsymbol{\alpha}) \le \mathbf{b}_{lifting}^{UB} \tag{17}$$

The constraint function vector $\mathbf{b}_{lifting}$ for this phase includes some of those in $\mathbf{b}_{BSB}$ that are imposed at all times: the system and contact constraints; foot configurations (positions and orientations, including *sl*) fixed to the desired; null undesired contact wrenches; and identical arm joint angles and a restricted range of the identical arm/object orientations.

Additional task requirements include constant upright neck joint angle ($q_{neck}(t) = 0,\ \forall t \in [0, T_1]$) and specific initial, intermediate, and final conditions. The initial conditions correspond to a given squat pose $\mathbf{q}^{squat}$ at rest, and the intermediate conditions for the standing pose are imposed on the CoM and the system pose as *desired*:

$$\mathbf{q}(0) = \mathbf{q}^{squat};\quad \dot{\mathbf{q}}(0) = \ddot{\mathbf{q}}(0) = \mathbf{0} \tag{18}$$

$$\bar{\mathbf{r}}(T_1) = \bar{\mathbf{r}}^{desired};\quad \dot{\bar{r}}_y(T_1) = \ddot{\bar{r}}_y(T_1) = 0 \tag{19}$$

$$q_{1,arm}(T_1) = q_{2,arm}(T_1) = q_{arm}^{desired};$$
$$\Phi_{1,arm}(T_1) = \Phi_{2,arm}(T_1) = \Phi_{arm}^{desired} \tag{20}$$

For continuity, $\mathbf{q}(T_1)$ and $\dot{\mathbf{q}}(T_1)$ from the two phases are required to be identical (Figure 3). As for the final conditions, the static equilibrium is imposed on both the CoM and the joints:

$$\dot{\bar{\mathbf{r}}}(T_2) = \ddot{\bar{\mathbf{r}}}(T_2) = \mathbf{0};\quad \dot{\mathbf{q}}(T_2) = \ddot{\mathbf{q}}(T_2) = \mathbf{0} \tag{21}$$

Additionally, for the lift-and-hold, the final CoM *X*-position is constrained to be within the BoS domain [–*fl*/2, *fl*/2+*sl*], and the CoM *Y*-position is lower-bounded to avoid an over-crouched pose:

$$-fl/2 \le \bar{r}_x(T_2) \le fl/2 + sl;\quad \bar{r}_y(T_2) \ge \bar{r}_y^{LB} \tag{22}$$

For the lift-and-release, the final pose is constrained to be identical to the initial squat pose:

$$\mathbf{q}(T_2) = \mathbf{q}(0) = \mathbf{q}^{squat} \tag{23}$$

For both tasks, any infeasible solution indicates that the given lifting task cannot be performed or the robot falls.

Fall prevention control requires a stability threshold. According to the sufficient conditions for imposing balanced states, the basin formed by a linear approximation of a pre-computed BSB corresponding to each task is imposed on the CoM state of the standing pose as an additional intermediate condition (Figure 3).

- Lift-and-hold: $(\bar{r}_x(T_1), \dot{\bar{r}}_x(T_1)) \in BSB_{(m_{object})}$ (24)
- Lift-and-release: $(\bar{r}_{robot,x}(T_1), \dot{\bar{r}}_{robot,x}(T_1)) \in BSB_{(m_{object}=0)}$ (25)

Here, $\mathbf{q}^{sampled}$ for the BSB construction is derived to be consistent with the above intermediate conditions for the CoM and system pose, ensuring the validity of imposing the BSB constraints. Consequently, it is $\dot{\bar{r}}_x(T_1)$ and $\dot{\bar{r}}_{robot,x}(T_1)$ that the BSB constraints are effectively imposed on in the trajectory optimization, through which, the system's full capability may be exploited.

## 7. Numerical Results and Comparative Analyses: Trade-off of Object Mass in Balancing Capabilities

The BSBs and the rates of change of momentum were computed for the reduced-order mechanism and humanoid robot with various object masses under different conditions. From the perspectives of BSBs as a measure of balancing capabilities, the associated trade-off relationships are analyzed.

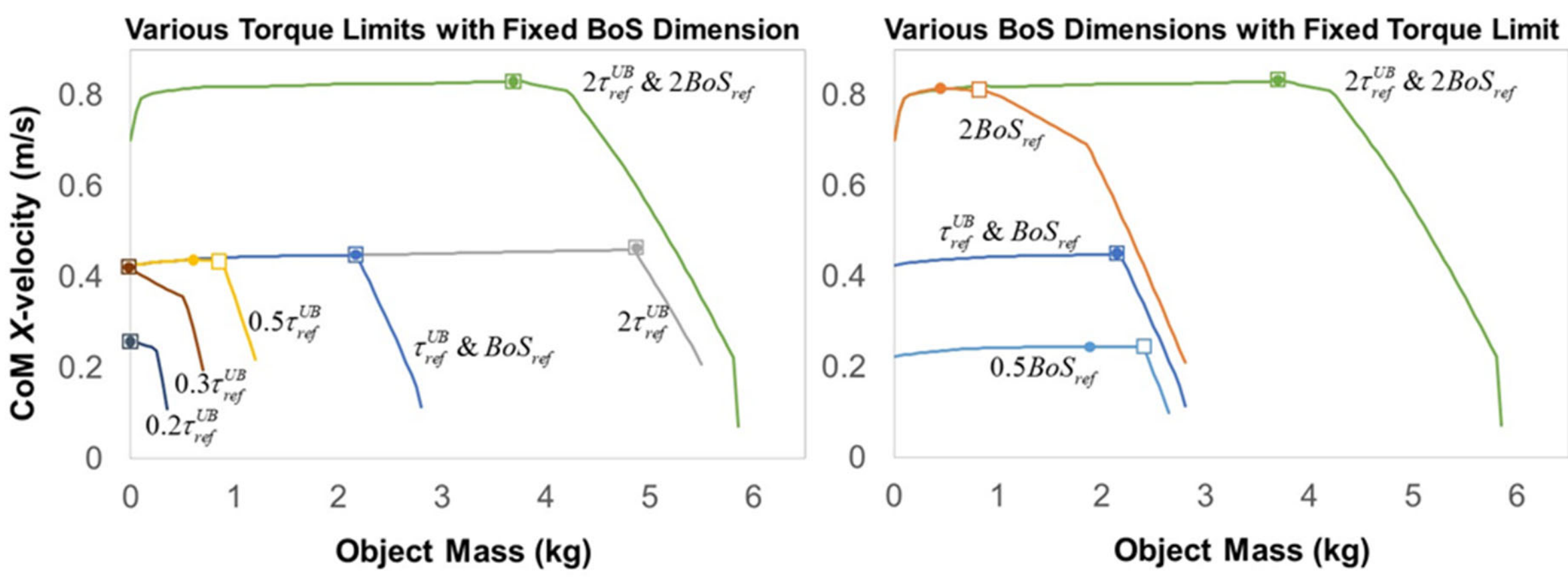

**Figure 4.** Balancing capabilities represented by the computed BSBs of the reduced-order mechanism versus object mass. For each condition, the CoM *X*-velocity is maximum at the critical mass (circle) and the joint torque limit becomes active at the transition mass (square) of the object.

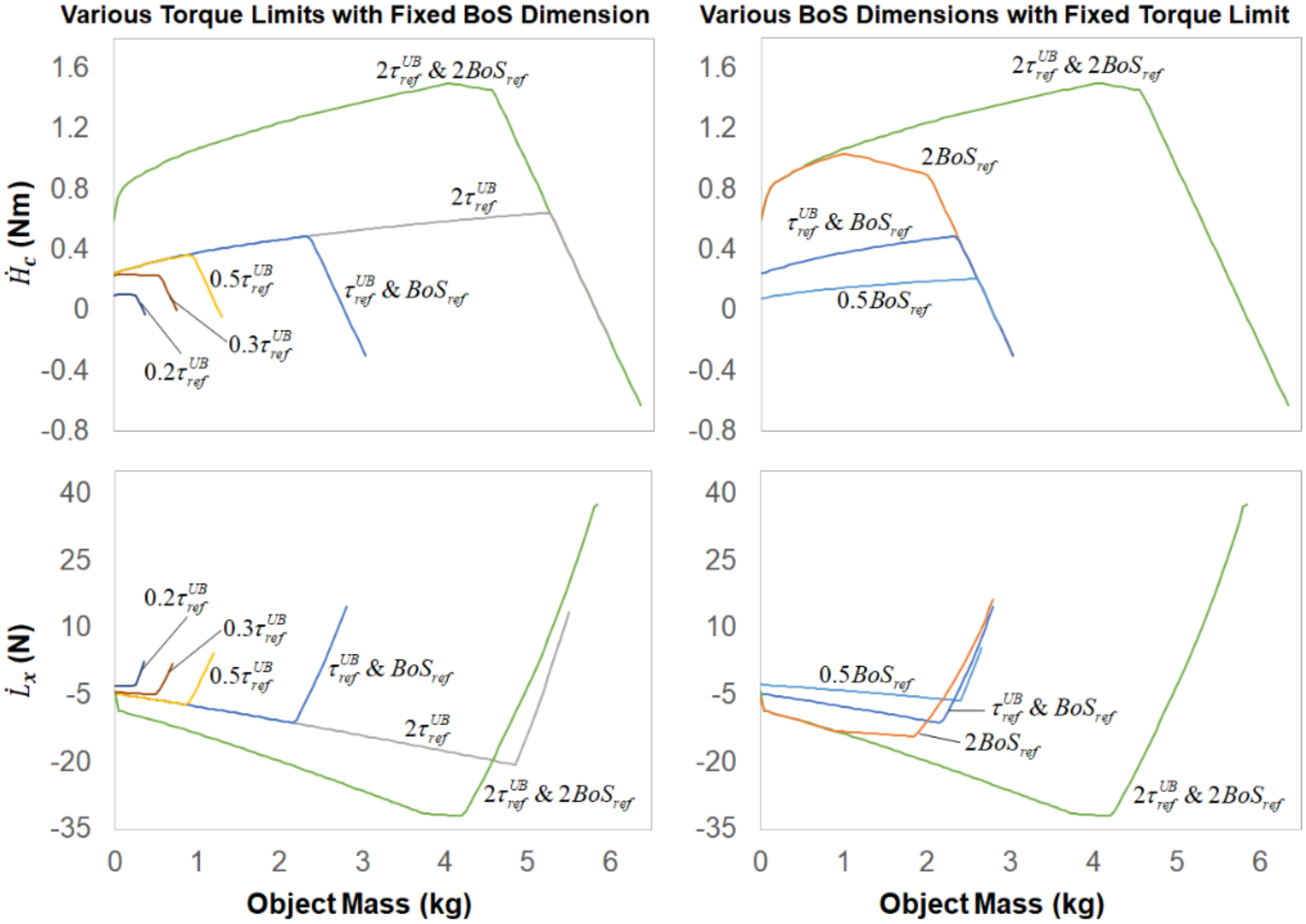

**Figure 5.** Computed centroidal angular (top) and linear (bottom) momentum rates of the reduced-order mechanism at the onset versus object mass

## *7.1 Reduced-Order Mechanism: Actuation Capacity and BoS*

For the reduced-order mechanism, the masses of the humanoid's torso and arm were assigned for the link mass $m_1$ and $m_2$, respectively. The link length $l_1 = 0.2$ m was used to represent the near-average CoM height of the humanoid robot during a lifting task and $l_2 = 0.13$ m was adopted from the robot's fully stretched total arm length. The object mass was varied from 0 to 5.85 kg in increments of 0.05 kg, beyond which the optimization problems were infeasible, indicating failure of maintaining balance. To evaluate the effects of the actuation capacity and BoS dimension on balancing capability, $\tau_{ref}^{UB}$ was scaled with 0.2–2.0 and $BoS_{ref}$ was scaled with 0.5, 1.0, and 2.0. The terminal time $T = 0.8$ s was tuned based on numerical experiments. For the final position, the CoM $X$-position range that corresponds to $0.5BoS_{ref}$ was used.

The system's balancing capability improved with increased object mass (as in the use of a balancing pole during tightrope or slackline walking), but only up to the critical mass when the CoM $X$-velocity is maximum, and deteriorates beyond, indicating the presence of trade-offs for stability (Figure 4). This is consistent with the system's capability to regulate its rates of momentum $\dot{L}_x(0)$ and $\dot{H}_c(0)$, which are maximum in magnitudes near the critical mass (Figure 5); they differ only by their signs as the system balances by moving its CoM position backward in response to the forward perturbations, resulting in the positive $\dot{H}_c(0)$ and negative $\dot{L}_x(0)$. This regulation is limited by the given actuation capacity and BoS dimension, which have contrasting effects on the balancing capability and the feasible domain of the object mass, as detailed below.

The active constraints for the torque and CoP from the optimization solutions (Table I) confirm the trade-off relationships across the domain of $m_{object}$ and numerically verify the analytical models for the transition mass. Considering $BoS = BoS_{ref} = fl$ with $m_1 = 1.74$ kg and $m_2 = 0.23$ kg, the joint torque limit required for the CoP constraint to be active for any $m_{object} < m_{obj,trans}$ is $\tau^{UB} > 0.5BoS_x(m_1 + m_2)g = 1.226 \simeq 0.344\tau_{ref}^{UB}$. This was confirmed by the numerical results of the BSBs with $0.2\tau_{ref}^{UB}$ and $0.3\tau_{ref}^{UB}$, where the CoP constraints were inactive and the torque limits were active for all feasible object masses (Figure 4, left). This implies that the balancing capability with $BoS_{ref}$ improves as the actuation capacity increases, but only up to this level.

With a sufficiently large torque limit ($\geq 0.344\tau_{ref}^{UB}$), for $m_{object} < m_{obj,trans}$, the CoP constraint is active while the torque limit is not (Table I), indicating that the BoS is the limiting factor and is fully exploitable for balancing. On the other hand, both the CoP and torque constraints were active at $m_{object} = m_{obj,trans}$, which was equal to or slightly greater than the critical mass. As predicted by the analytical model for $m_{object} = m_{obj,trans}$, the joint torque limit constraint becomes active with a smaller object mass (transition mass) as the BoS dimension increases (Figure 4, right and Table I). For $m_{object} > m_{obj,trans}$, the torque limit constraint is active while the CoP constraint is not, indicating that the actuation capacity is the limiting factor and is fully exploitable for balancing. In this domain, the BSBs decrease rapidly as $m_{object}$ increases. The decreasing $\dot{H}_c(0)$ and increasing $\dot{L}_x(0)$ as $m_{object}$ increases beyond $m_{obj,trans}$ (Figure 5) are the results of the system's balancing effort with a large object mass under the actuation capacity as the limiting factor.

The results also reveal the contrasting effects of the actuation capacity and BoS dimension on the balancing capability. With the joint torque limit fixed, the increase in the BoS dimension increases the BSB for a given $m_{object}$, but has almost no effect on the feasible domain of $m_{object}$ (Figure 4, right and Table I). These attributes of the BSBs are in agreement with those of the momentum rates of the system (Figure 5). This again implies that a sufficiently large BoS is required to fully utilize the available joint actuation required for balancing.

In contrast, the joint torque limits that are sufficiently large ($\geq 0.344\tau_{ref}^{UB}$) do not affect the balancing capability, resulting in almost identical BSBs for a given BoS dimension and $m_{object}$ below their respective critical mass (Figure 4, left and Table I). This saturation of BSBs is consistent with the prior sensitivity analysis results for an inverted pendulum model without an object (Gonzalez et al., 2018). On the other hand, increases in joint torque limit extend the feasible domain of $m_{object}$. This domain is upper-bounded approximately by the maximum object mass that the system can sustain at the given initial pose with the required torque $\tau_2 \geq (m_2 + m_{object})gl_2$ at the second joint. For a given torque limit, the maximum $m_{object}$ derived under this assumption is near and slightly smaller than the maximum feasible $m_{object}$ obtained from the numerical optimization—for example, 2.6 kg and 2.8 kg, respectively, for the condition with $\tau_{ref}^{UB}$ and $BoS_{ref}$ (Table I and Figures 4 and 5).

## *7.2 Humanoid Robot: BoS and CoM Height*

The results were computed with $m_{object}$ ranging from 0 to 1.5 kg (42.86% of $m_{robot}$) in the state space of CoM *Y*-position and CoM *X*-velocity (Figure 6). This state space represents the robot's varying pose (CoM height), such as in object lifting, under fore-aft perturbations. The initial poses $\mathbf{q}^{sampled}$ were generated as sampled from the ZMP-controlled quasi-static lifting motions in a prior work (Song et al., 2023a) such that the total (i.e., including both the robot and the object) CoM $X$-position is at the center of its BoS, both arms are held forward and parallel to the ground ($q_{arm}^{desired} = \pi/2$ and $\Phi_{arm}^{desired} = 0$), and the torso is upright. To avoid self-collision, the orientations of both arms with the object were restricted by $-\pi/4 \le \Phi_{1,arm}(t) = \Phi_{2,arm}(t) \le \pi/2$, $\forall t \in [0,T]$. Two BoS dimensions, $fl$ (with $sl = 0$) and 1.5$fl$ (with $sl = 0.5fl$), were considered on the level ground. For the final position, the CoM $X$-position range that approximately corresponds to $fl$ was used for both BoS dimensions. The CoM $Y$-positions were sampled from the hip joint height varying from 0.12 m to 0.19 m in 21 equally-spaced values. The combinations of the sampled CoM positions and BoS dimensions also result in various robot poses and the associated object positions. The terminal time $T$ was set to 0.95 s for $sl = 0$ and 1.0 s for $sl = 0.5fl$ based on numerical experiments. The computed BSBs and the rates of change of angular momentum were approximated with third- to fifth-order polynomials to reduce numerical noise without overfitting the data.

| Torque limit | BoS dimension | $m_{obj,trans}$ (kg) | Active Constraint(s) | | | Max. $\dot{r}_x(0)$ (m/s) | Max. feasible $m_{object}$ (kg) |
|---|---|---|---|---|---|---|---|
| | | | $m_{object} < m_{obj,trans}$ | $m_{object} = m_{obj,trans}$ | $m_{object} > m_{obj,trans}$ | | |
| $0.2\tau_{ref}^{UB}$ | $BoS_{ref}$ | 0 | N/A | Torque | Torque | 0.26 | 0.35 |
| $0.3\tau_{ref}^{UB}$ | | | | | | 0.42 | 0.7 |
| $0.5\tau_{ref}^{UB}$ | | 0.9 | CoP | Torque & CoP | | 0.44 | 1.2 |
| $\tau_{ref}^{UB}$ | | 2.15 | | | | 0.45 | 2.8 |
| $2\tau_{ref}^{UB}$ | | 4.85 | | | | 0.46 | 5.5 |
| $\tau_{ref}^{UB}$ | $0.5BoS_{ref}$ | 2.4 | | | | 0.24 | 2.65 |
| | $BoS_{ref}$ | 2.15 | | | | 0.45 | 2.8 |
| | $2BoS_{ref}$ | 0.9 | | | | 0.81 | 2.8 |
| $2\tau_{ref}^{UB}$ | $2BoS_{ref}$ | 3.75 | | | | 0.83 | 5.85 |

**Table I.** Numerical results with torque limits and BoS dimensions for the reduced-order mechanism

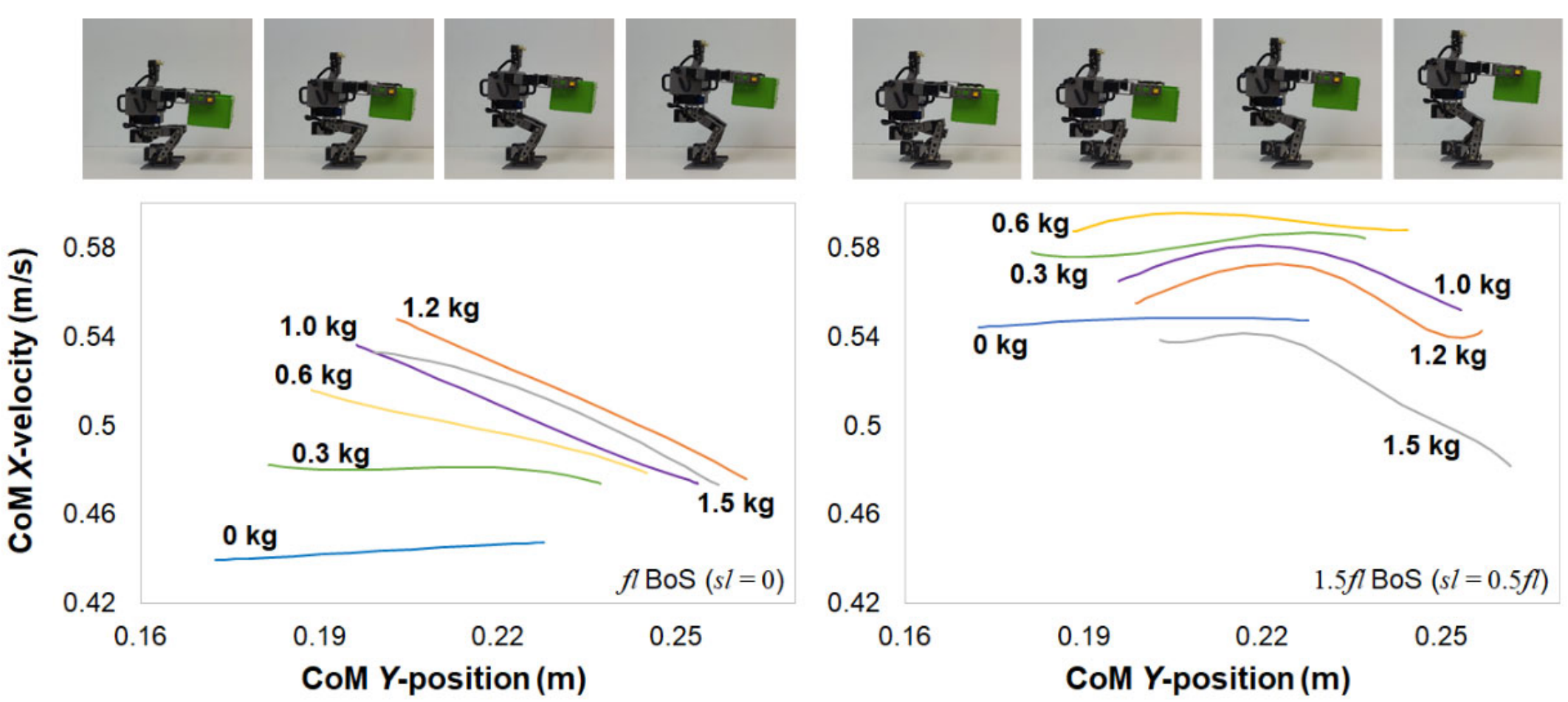

**Figure 6.** Balancing capabilities: The computed BSBs versus CoM heights (bottom) from the sampled poses of the humanoid robot (top) holding various object masses with two different step lengths.

The trade-off relationships were also present for the humanoid robot. As the object mass increases, the BSB increases and then decreases. The critical mass was 1.2 kg for $fl$ BoS and 0.6 kg for 1.5$fl$ BoS, which is approximately 34.28% and 17.14%, respectively, of $m_{robot}$. In addition, for a given $m_{object}$, the increase of the BoS dimension increased the balancing capability but this effect diminishes as $m_{object}$ increases further (Figures 6 and 7). The overall trends in the critical masses and BSBs of the humanoid are similar to those of the reduced-order mechanism. The ranges of BSBs for the humanoid are also within those of the reduced-order mechanism with $\tau_{ref}^{UB}$ and $BoS_{ref}$ across the common domain of $m_{object}$ (Figures 4 and 6), while the inclusion of the whole-body dynamics and properties resulted in some differences, particularly for large object masses. The distinct results between the two step lengths were made available by the distribution formulation for the contact wrenches and CoPs that reflect the dynamic effects of the object mass.

The results also elucidate the novel CoM-state space analyses with horizontal velocity perturbations for different vertical CoM heights. For the $fl$ BoS, most BSBs have negative slopes with increased magnitudes as $m_{object}$ increases, indicating reduced balancing capabilities for higher CoM $Y$-positions (Figure 6). This trend is consistent with the results of another stability model based on a linear inverted pendulum (Koolen et al., 2012), in which the allowable perturbation velocity was smaller with a higher CoM position for a given capture point.

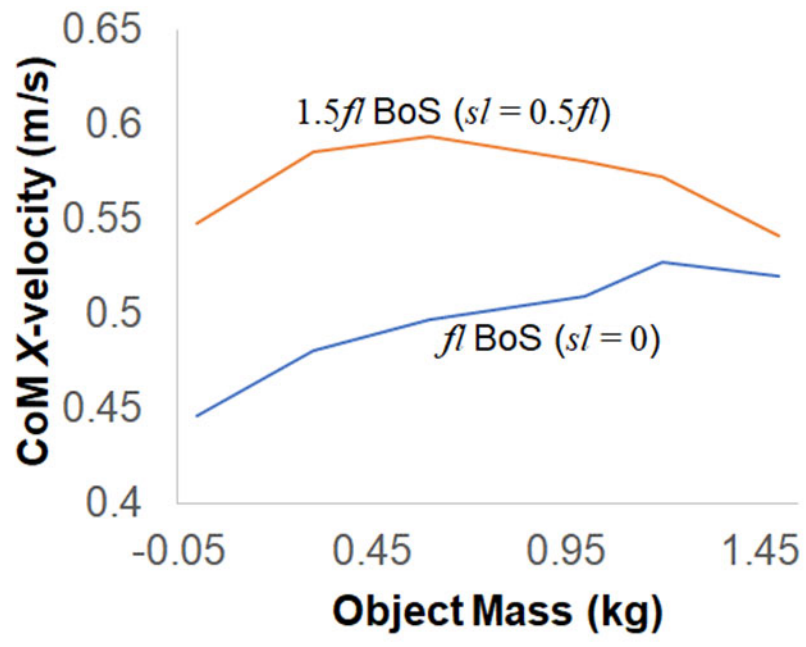

**Figure 7.** Computed BSBs of the humanoid robot versus object mass at a sample CoM $Y$-position of 0.22 m

The increase in the magnitude of the rates of change of angular momentum as $m_{object}$ increases (Figure 8, top) is consistent with the results of the reduced-order mechanism, except for those beyond the critical mass, which is likely due to the complexity in the humanoid model such as large DoFs, high redundancy, and restricted joint variables. For the 1.5*fl* BoS, the BSBs are almost the same with the increased CoM *Y*-position up to the critical mass (0.6 kg), beyond which they are concave. With the non-zero step length when both legs are asymmetric, inverted-pendulum-based analysis does not apply, and the nonlinear attributes of the BSBs are likely due to the combinations of several factors, such as different required joint torques and reduced joint ranges of motion between the two legs. These also resulted in the nonlinearity in the rates of change of angular momentum of the system with 1.5*fl* BoS, in contrast to the relatively ordered results with *fl* BoS (Figure 8).

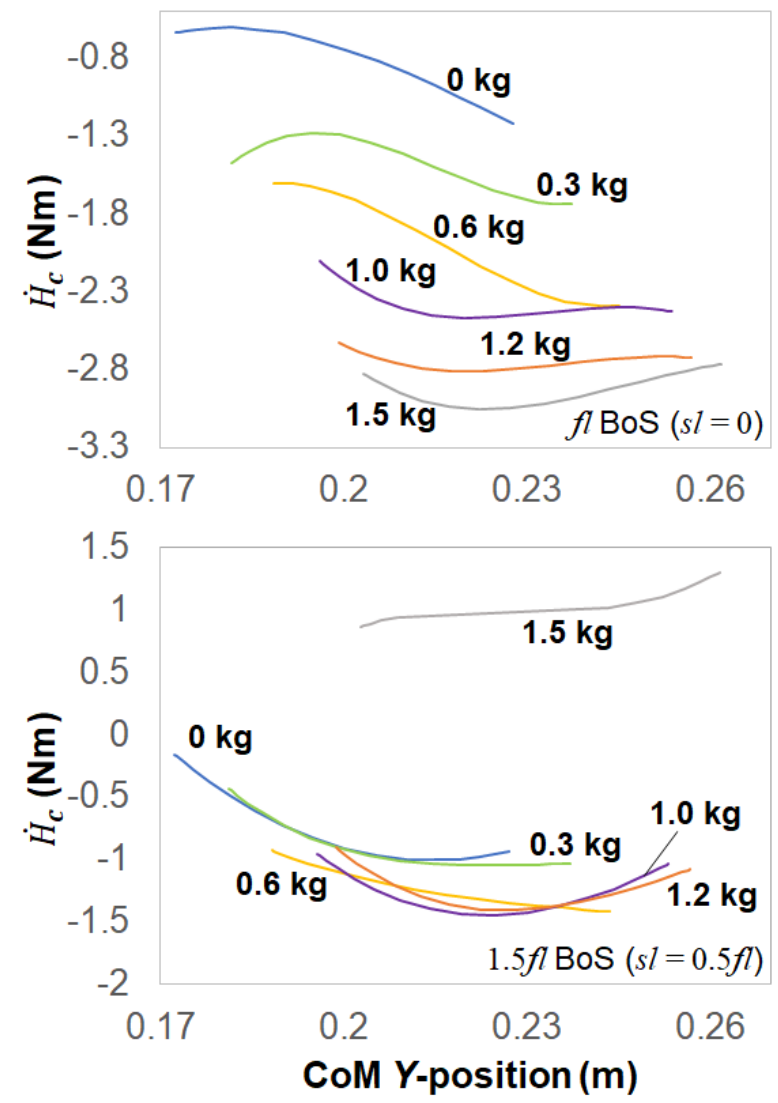

**Figure 8.** Computed centroidal angular momentum rates of the humanoid robot at the onset versus CoM *Y*-position with various object masses and two different step lengths

The BSBs with $sl = 0$ and the object mass of 0 kg (unloaded) and 0.6 kg (half of the critical mass) were computed also for the CoM-state space of *X*-position and *X*-velocity (Figure 9; see also Extension 1). The initial poses were generated such that the hip joint height is at 0.19 m and the CoM *X*-position is in the selected range from –0.053 m to 0.051 m within *fl* BoS. In consistency with the above results, $BSB_{(0.6\text{ kg})}$ is larger than $BSB_{(0\text{ kg})}$, confirming the benefit of the object mass, below the critical mass, in balancing.

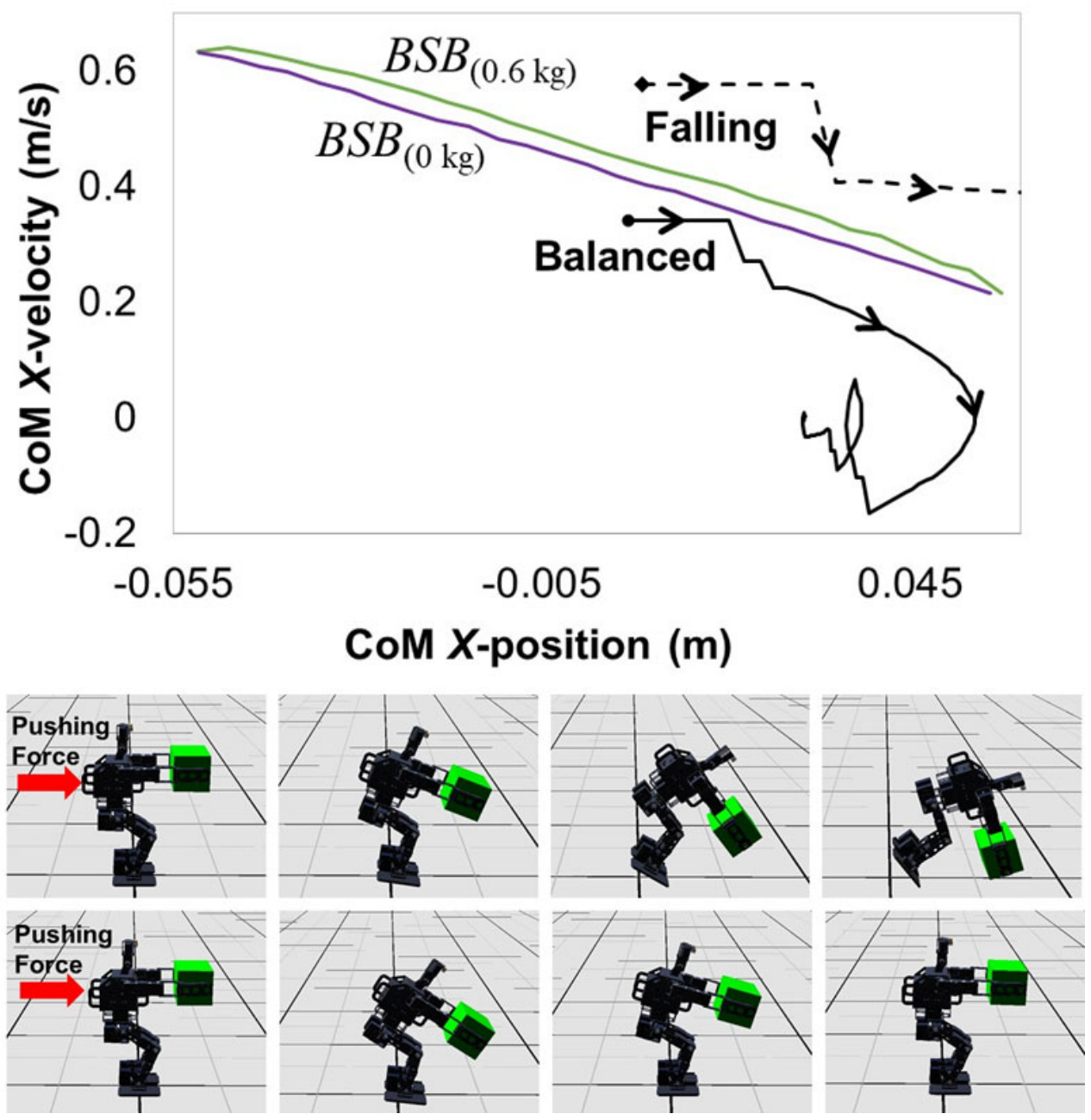

**Figure 9.** Simulation validations: The BSBs versus CoM *X*-position for the object masses of 0 kg and 0.6 kg with a given hip joint height and the balanced and falling CoM-state trajectories. The snapshots of the simulated motions correspond to the balanced (bottom) and falling (top) trajectories.

For the illustrative validation of the computed BSBs, an existing hip-strategy-based balance controller from prior works (Peng et al., 2022; Song et al., 2024) was implemented for push-recovery tests in a simulation environment (Webots, Cyberbotics Ltd.). The parameters and gains of the controller were manually tuned. Two force perturbations were applied to the robot with $m_{object} = 0.6$ kg. When the initial CoM state was inside $BSB_{(0.6\text{ kg})}$, the resulting CoM-state trajectory remained within and the robot was able to maintain balance (Figure 9). When the initial CoM state was outside $BSB_{(0.6\text{ kg})}$, the CoM-state trajectory remained outside, resulting in a fall. Despite a few discrepancies in the conditions between the BSB formulation and the simulation environment (e.g., foot tipping), which also have offsetting effects (Peng et al., 2022), these simulated results and comparison demonstrate the validity of BSBs as accurate stability thresholds.

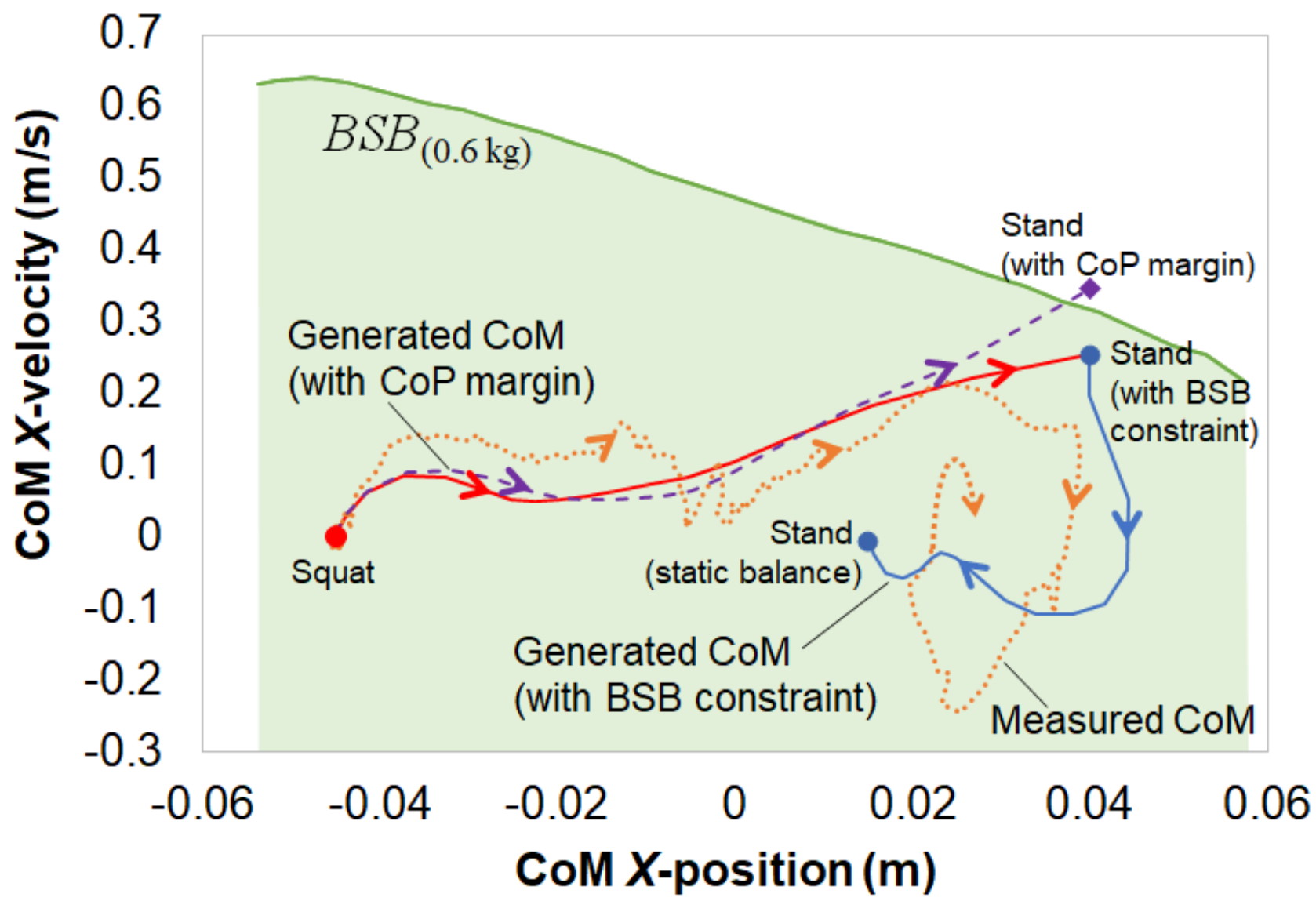

**Figure 10.** The computed $BSB_{(0.6\ \mathrm{kg})}$ (shaded area) and the CoM-state trajectories for the lift-and-hold task. The generated squat-to-stand phase with the BSB constraint (solid) reached the standing state inside $BSB_{(0.6\ \mathrm{kg})}$, while that with the CoP margin (dashed) exited it. The generated stabilization phase (solid) from standing with the BSB constraint reached static balance at the final state. The measured trajectory (dotted) from the experiment (omitting some vibrating portion due to tipping and tracking errors after achieving the desired pose) follows approximately the one generated with the BSB constraint.

## 8. Simulation and Experimental Demonstration Results: Object Lifting of Humanoid Robot While Balancing

All lifting tasks were demonstrated on level ground with $sl = 0$ m and $m_{object} = 0.6$ kg (half of the critical mass). The bounds on the arm/object orientation, $-\pi/4 \le \Phi_{1,arm}(t) = \Phi_{2,arm}(t) \le \pi/2$, $\forall t \in [0, T]$, for self-collision avoidance in the BSB construction, were also used here. The solution trajectory was tracked using a proportional-integral-derivative (PID) control at the servomotors, complemented by a proportional adjustment, $\mathbf{q}^{updated} = \mathbf{q}^{desired} + K(\mathbf{q}^{desired} - \mathbf{q}^{measured})$, where $K$ is a diagonal matrix of proportional gains. The adjustment was implemented on the robot's main computer to enhance trajectory tracking performance in experiments. At each time step, the *measured* encoder values were used to obtain the *updated* control inputs to the servomotor for the next time step based on the *desired* joint variables (whole-body pose) from the generated trajectory. All control gains were tuned to reduce the tracking error. The data from the encoders and inertial measurement unit into forward kinematics and a moving average filter with a window size of 2 and 5 were used to estimate the CoM positions and velocities, respectively.

### *8.1 Lift-and-Hold*

The initial squat pose $\mathbf{q}^{squat}$ for the lift-and-hold was assigned with a total CoM $X$-position of –0.045 m, an upright torso with the hip joints at a height of 0.12 m, and the arms forward and parallel to the ground ($q_{arm}^{desired} = \pi/2$ and $\Phi_{arm}^{desired} = 0$). For the intermediate standing pose at $t = T_1$, $(\bar{r}_x^{desired}, \bar{r}_y^{desired}) = (0.04,\ 0.245)$, where $\bar{r}_y^{desired}$ corresponds to the hip joint height of 0.19 m used for the BSB construction, and the arms are the same as in the initial pose. Other parameters were set as $T_1 = 1.0$ s, $T_2 = 1.64$ s, and $\bar{r}_y^{LB} = 0.21$ m. For comparative evaluation against a baseline criterion, another lift-and-hold trajectory was generated with the BSB constraint replaced by a margined range for CoP (Dafarra et al., 2020; Orin et al., 2013), equivalent to ZMP for the case of no relative motion at the contact between the foot and the ground, which is commonly used for stable robot control (Feng et al., 2014; Kajita et al., 2001; Murooka, Ueda, et al., 2017; Shigematsu et al., 2019). The margined CoP range imposed was $-0.058 \le {}^{p}r_{CoP,x} \le 0.058$, corresponding to 91.34% of each $p$-th BoS dimension, during the squat-to-stand phase. The solution value of the cost function for the squat-to-stand phase with the BSB constraint was 4.481 $N^2m^2s$ and that with the CoP margin was 4.784 $N^2m^2s$.

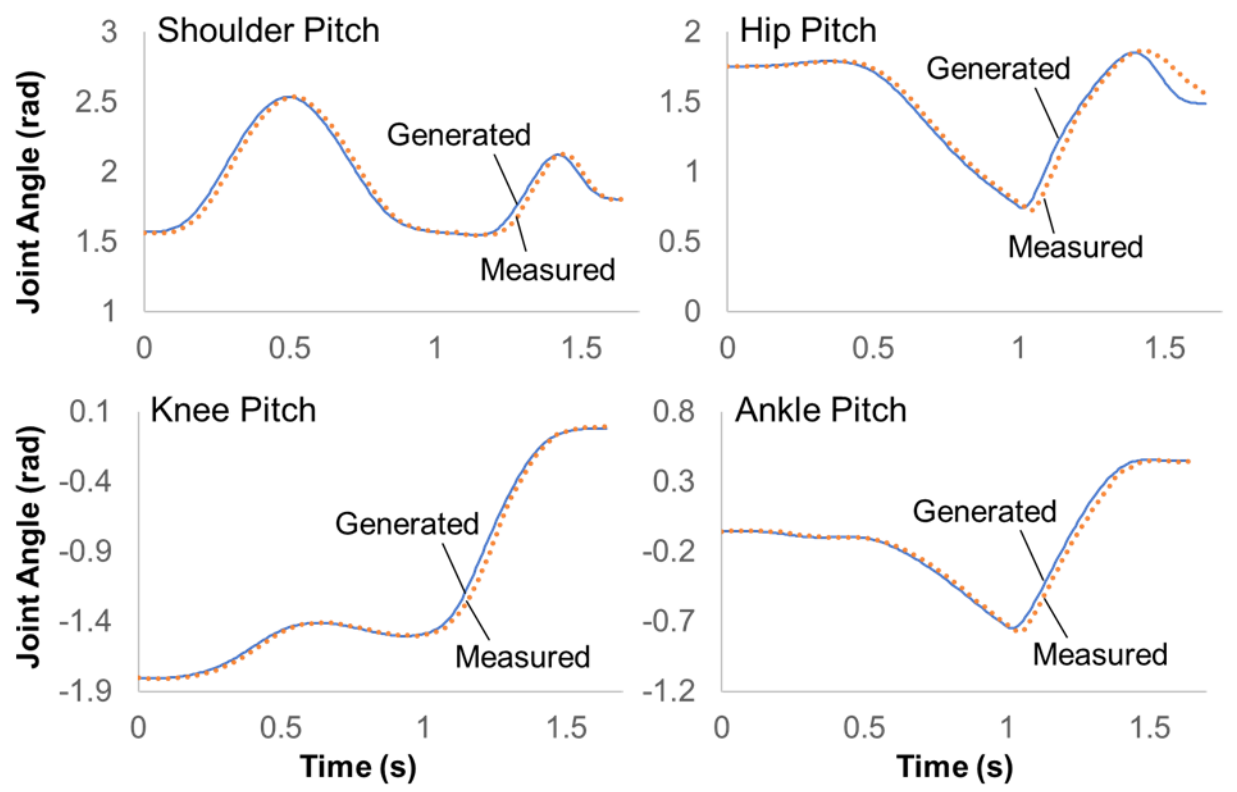

**Figure 11.** Joint variable trajectories for the lift-and-hold task with the BSB constraint: generated (solid) and experimentally measured (dotted)

The two thresholds show significant differences (Figure 10). For the trajectory with the CoP margin, the standing state at $t = T_1$ exited $BSB_{(0.6\text{ kg})}$. According to the definition, final static balance could not be achieved from this state, which was also proven by the infeasible solution of the optimization problem for the subsequent stabilization phase where the intermediate conditions of joint positions and velocities were violated. This also affirms that the balancing capability is limited not only by the system/contact constraints but also by the current joint states. Furthermore, the CoP margin constraint was never active and the resulting CoP ranged from –0.44 m to –0.019 m, which were well within its imposed bounds as well as the entire BoS dimension. This illustrates that the CoP or ZMP bound itself cannot be an accurate balance stability threshold even when used with a sufficiently large margin. In contrast, the imposition of the BSB threshold as a constraint ensured that the standing state of lifting remains inside the basin, leading to a successful solution trajectory for the subsequent stabilization phase toward final static balance. The complete solution trajectory closely matched overall with that from the experimental measurement (Figure 10), except for some deviation due to a degree of foot tipping after achieving the desired final pose.

The results from the trajectory optimization were confirmed in both simulations and experiments, and the trajectory tracking performance was verified (Figure 11). The humanoid robot successfully performed the BSB-constrained lifting task without falling, while it fell down with the CoP margin (Figure 12; see also Extension 2), consistent with the solutions of the trajectory optimization. A degree of foot tipping observed in the simulations and experiments is an unavoidable source of deviations from the generated trajectories with the BSB constraints that are computed under full contact conditions. Nevertheless, the foot tipping was minor and momentary, confirming the validity of the comparative analyses. In agreement with the CoM-state trajectory with the BSB constraint that remained within $BSB_{(0.6\text{ kg})}$ (Figure 10), the robot successfully achieved final static balance. In particular, at the end of the squat-to-stand phase, the CoM state approached near the boundary of $BSB_{(0.6\text{ kg})}$, indicating that the lifting trajectory generation fully utilized the robot's balancing capability. Overall, these results and comparisons validate the accuracy of BSB as a stability threshold and demonstrate the merit of imposing it as a constraint for trajectories.

### *8.2 Lift-and-Release*

The initial squat pose $\mathbf{q}^{squat}$ for the lift-and-release was assigned with a total CoM $X$-position of 0 m, an upright torso with the hip joints at a height of 0.12 m, and the arms forward and parallel to the ground ( $q_{arm}^{desired} = \pi / 2$ and $\Phi_{arm}^{desired} = 0$ ). For the intermediate standing pose, $T_1$ = 1.5 s

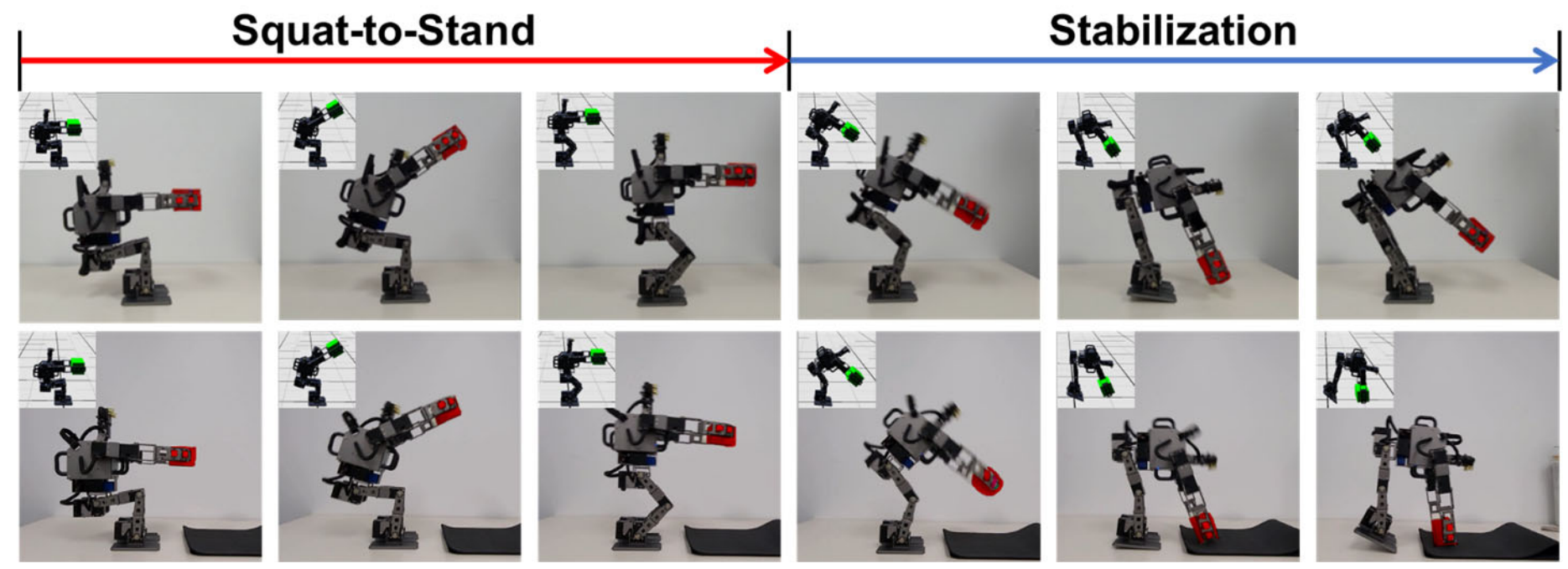

**Figure 12.** Snapshots of the lift-and-hold trajectory experiments and simulations (insets) with the BSB constraint (top) and with the CoP margin (bottom)

and $(\overline{r}_x^{desired}, \overline{r}_y^{desired}) = (0.058,\ 0.245)$ with $\overline{r}_y^{desired}$ corresponding to that used for the BSB construction, and the arms are the same as in the initial pose. For the final squat pose, $T_2$ = 2.4 s. The solution value of the cost function for the squat-to-stand phase was 6.036 $N^2m^2s$.

The robot's CoM-state trajectory measured from the simulation matched closely with that generated from the solution, except briefly during the stabilization phase due to some degree of foot tipping (Figure 13; see also Extension 3). During the squat-to-stand phase, the total CoM state approached the boundary of $BSB_{(0.6\ \text{kg})}$ prior to the object release, an indication of full utilization of its balancing capability. During the stabilization phase, the robot's CoM-state trajectory remained well inside $BSB_{(0\ \text{kg})}$, as its *X*-position shifted backward discontinuously upon object release. As a result, the robot successfully achieved static balance at the final squat pose, identical to its initial pose, for continuous operation without falling. This result demonstrated that the challenge of balance stability problem in trajectory optimization when considering discontinuous mass properties can be addressed by using the BSB corresponding to the changed mass properties as a stability threshold in a desired contact for the state at the instant of the change.

## 9. Conclusions and Future Work

The object mass parameters were incorporated into BSB through the whole-body dynamics with stance contacts, which provided systematic quantification of their dynamic effects on balancing. The reduced-order mechanism allowed for rigorous cause-and-effect evaluations and analytical insights into the BSB results for the humanoid robot. In particular, the computed BSBs confirmed the analytical prediction of the trade-offs between the momentum regulations and limiting factors—including the contrasting effects of the actuation capacities and BoS dimensions—with respect to the object mass in a system's balancing capabilities. The trade-offs generally resulted in the maximum balancing capability at the critical mass of the object, while different limiting factors were activated at the transition mass. The sufficient conditions for imposing BSBs as the threshold constraints in the trajectory optimization were demonstrated in the humanoid robot control for the lift-and-hold and lift-and-release tasks that have distinct mass properties, and was validated in simulations and experiments. The merits of BSB in accurately indicating balance stability were also illustrated against the existing criterion of CoP (or ZMP) margin. In real-world applications of the proposed framework, the unknown object mass parameters can be estimated using established methods (Murooka, Nozawa, et al., 2017; Shigematsu et al., 2018; Han et al., 2020). Overall, the results highlight the implications of BSB as a measure in predictions and as a threshold in controls regarding the dynamic effects of an object mass on balance stability.

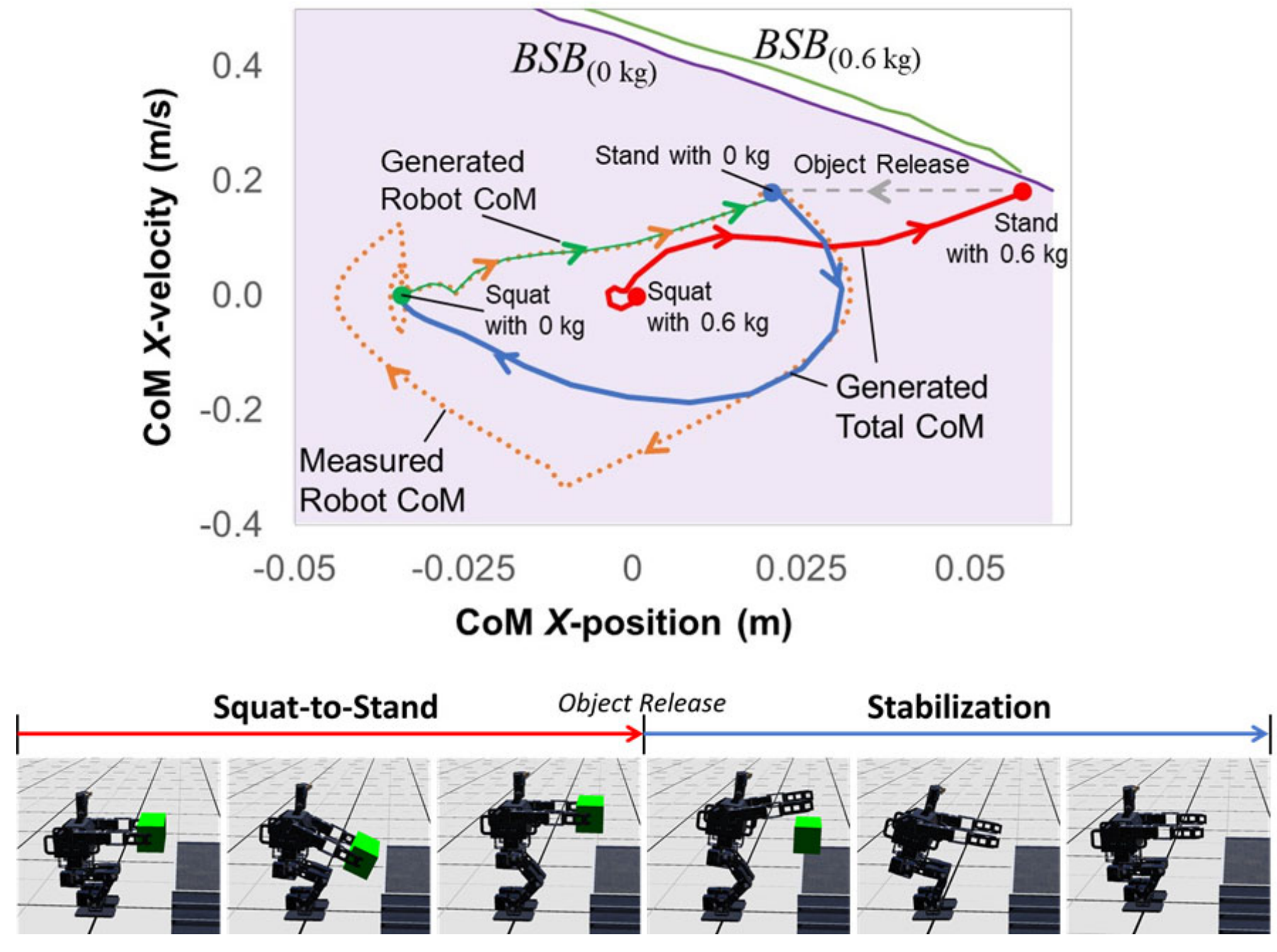

**Figure 13.** The computed $BSB_{(0\ \text{kg})}$ (shaded area) and $BSB_{(0.6\ \text{kg})}$ and the CoM-state trajectories for the lift-and-release task. The generated total CoM-state trajectories (thick-solid) include the squat-to-stand and stabilization phases, connected by a sudden shift (dashed) due to the release of the object. The robot (i.e., without the object mass) CoM-state trajectory measured from the simulation (dotted) follows approximately the ones generated for the squat-to-stand (thin-solid) and stabilization (thick-solid) phases, and they all reached static balance at the final state. The snapshots from the trajectory simulation are shown at the bottom where the released object is moved by a conveyor belt.

Some limitations of the current approach prompt further exploration. The computational cost of the optimization may be addressed by offline computation and storage for real-time access by an online controller and by exploiting the linear dynamics of object mass through parameter continuation (Li and Dankowicz, 2020). While the demonstration examples in this study were confined to the 2D sagittal plane in DS for their analytical and computational tractability and visualization of their cause-and-effect relationships, the proposed framework is general, and can be implemented under perturbations at various positions (e.g., joints, slips, and trips) in any 3D directions, which may involve stepping (Pratt et al., 2006; Koolen et al., 2012; Peng et al., 2022; Song et al., 2024; Bodmer et al., 2026). In addition, future work will extend and demonstrate the framework with multiple humanoid platforms and their controller-specific properties in various poses for benchmarking. The overall approach can be advanced to inform the design and control of humanoid robots for stable loco-manipulation tasks.

**Acknowledgements**

Portions of this work have been adapted, extended, and generalized from preliminary results in the Ph.D. dissertation of the first author and a presentation at the 22nd IEEE-RAS International Conference on Humanoid Robots (Austin, TX, USA, 2023).

**Declaration of Conflicting Interests**

The author(s) declared no potential conflicts of interest with respect to the research, authorship, and/or publication of this article.

**Funding Statement**

This work was supported in part by the U.S. National Science Foundation (IIS-1427193), New York University (NYU) School of Engineering Fellowships, and NYU Center for Urban Science and Progress (CUSP) Seed Grant.

**Appendix A: Index to Multimedia Extensions**

1 Video Validations of balanced and falling trajectories.

2 Video Lift-and-hold control experiments and simulations.

3 Video Lift-and-release control simulation with BSB constraint.